%% file: neurips_2025.tex
\documentclass{article}

\usepackage[final]{neurips_2025}

\usepackage[utf8]{inputenc} % allow utf-8 input
\usepackage[T1]{fontenc}    % use 8-bit T1 fonts
\usepackage{hyperref}       % hyperlinks
\usepackage{url}     
\usepackage{booktabs}       % professional-quality tables
\usepackage{amsfonts}       % blackboard math symbols
\usepackage{nicefrac}       % compact symbols for 1/2, etc.
\usepackage{microtype}      % microtypography
\usepackage{xcolor}         % colors
\definecolor{lightgray}{gray}{0.9}
\usepackage{adjustbox}

\newcommand{\meanstd}[2]{#1{\scriptsize$\pm$#2}}
\usepackage{multirow}
\usepackage{amsmath,amssymb}
\usepackage{algorithm}
\usepackage{algpseudocode}
\usepackage{bbm}
\usepackage{cleveref}
\usepackage{amsthm}

\newcommand{\ourmethod}{{O3SRL}\,}

\newtheorem{defn}{Definition}

\newtheorem{theorem}{Theorem}

\def\OfflineOracle{{\mathbb{O}_{\text{offline-RL}}}}
\def\StochasticOfflineOracle{{\mathbb{SO}_{\text{offline-RL}}}}
\def\epsoffline{{\epsilon_{\text{offline-RL}}}}
\def\noregret{{\text{NO\_REGRET\_UPDATE}}}

\title{Online Optimization for Offline Safe \\ Reinforcement Learning}

\author{
  Yassine Chemingui \\
  Washington State University \\
  \texttt{yassine.chemingui@wsu.edu} \\
  \And
  Aryan Deshwal \\
  University of Minnesota \\
  \texttt{adeshwal@umn.edu} \\
  \And
  Alan Fern \\
  Oregon State University \\
  \texttt{alan.fern@oregonstate.edu} \\
  \And
  Thanh Nguyen-Tang \\
  New Jersey Institute of Technology \\
  \texttt{thanh.nguyen@njit.edu } \\
  \And
  Janardhan Rao Doppa \\
  Washington State University \\
  \texttt{jana.doppa@wsu.edu} \\
}

\begin{document}

\maketitle

\begin{abstract}
 We study the problem of Offline Safe Reinforcement Learning (OSRL), where the goal is to learn a reward-maximizing policy from fixed data under a cumulative cost constraint. We propose a novel OSRL approach that frames the problem as a minimax objective and solves it by combining offline RL with online optimization algorithms. We prove the approximate optimality of this approach when integrated with an approximate offline RL oracle and no-regret online optimization. We also present a practical approximation that can be combined with any offline RL algorithm, eliminating the need for offline policy evaluation. Empirical results on the DSRL benchmark demonstrate that our method reliably enforces safety constraints under stringent cost budgets, while achieving high rewards. The code is available at \url{https://github.com/yassineCh/O3SRL}.

\end{abstract}

\input{introduction}

\input{problem-setup}

\input{related-work}

\input{technical-approach}

\input{experiments}

\input{summary}

\newpage
\bibliographystyle{abbrvnat}
\bibliography{osrl}

\newpage

\appendix
\input{appendix}
%%%%%%%%%%%%%%%%%%%%%%%%%%%%%%%%%%%%%%%%%%%%%%%%%%%%%%%%%%%%
\newpage

\end{document}

%% file: introduction.tex
\section{Introduction}

Offline reinforcement learning (RL) \citep{levine2020offline} is a powerful paradigm to learn decision-making policies from logged datasets without the need for additional interaction with the environment. Offline RL has shown good success in domains including autonomous driving, robotics, control systems, and black-box optimization \citep{autonomous1,autonomous2,robotics1,power1,PGS} where executing exploratory actions is not practical. However, in safety-critical domains such as healthcare and smart grid, the decision-making agent needs to also satisfy some cost/safety constraints. Such problems are studied under the sub-area referred to as offline safe RL (OSRL) where the goal is to learn reward-maximizing policies which satisfy cost constraints from offline datasets \citep{wachi2024survey}. 

OSRL inherits the challenges from both offline RL and safe RL \citep{wachi2024survey}. First, handling distributional shift when the learned policy encounters states and actions not seen in the offline dataset. Prior work in offline RL addresses this challenge by some form of penalization to realize the general principle of pessimism in the face of uncertainty \citep{levine2020offline}. Second, ensuring that the learned policy meets the cost/safety constraints after deployment. Maintaining the cost constraints typically requires off-policy evaluation (OPE) procedures which are highly unstable and typically have estimation errors \citep{offline_rl_survey_2}. 
OSRL is an active research topic and is relatively less-studied than offline RL. Constrained policy optimization using Lagrangian relaxation \citep{polosky2022constrained, lee2022coptidice,xu2022constraints} is a popular approach for OSRL, but due to the need to solve intertwined optimization problems they are unstable in practice, and are shown to have bad performance either causing oscillation/divergence or learning overly conservative (near zero-reward) policies. Additionally, OSRL problem with stringent safety constraints (i.e., small cost thresholds) is an important but severely under-studied problem \citep{zheng2024safe} and arises in many safety-critical control applications such as industrial process and energy system control.

This paper develops a novel offline constrained RL framework referred to as {\em {\bf O}nline {\bf O}ptimization for {\bf O}ffline {\bf S}afe {\bf RL} (\ourmethod)}. \ourmethod formulates a minimax optimization problem for policy learning and solves it using an iterative approach that relies on two key components, namely, an offline RL oracle and a no-regret algorithm. We perform two main algorithmic steps in each iteration of O3SRL. First, we produce a distribution over policies from solving an offline RL problem by defining a new reward function that combines the original reward and cost values based on the values of the current Lagrange variable and the cost threshold. Second, we employ a no-regret algorithm \citep{OCO} to adaptively update the value of Lagrange variable based on the current distribution over policies. We prove that the O3SRL framework is guaranteed to converge to the minimax solution. 

Unfortunately, the general O3SRL framework suffers from two drawbacks in practice. First, the use of no-regret algorithms over a continuous Lagrange variable requires calling OPE procedures which are typically unstable \citep{offline_rl_survey_2} and can potentially lead to error propagation over iterations. Calling an OPE procedure in each iteration also contributes to computational expense. Second, it is computationally-expensive to run an offline RL algorithm to convergence in each iteration. To overcome these challenges, we develop a practical and effective algorithm with convergence guarantees. We employ $K$ discrete values for Lagrange variable and apply a multi-armed bandit algorithm. Since most offline RL algorithms are based on stochastic gradient descent, we perform a small number of gradient updates on the policy from the previous iteration for efficiency. One advantage of this approach is that we can leverage powerful offline RL algorithms in a plug-and-play manner to solve offline constrained RL problems.

We perform experimental evaluation of this practical approach on multiple DSRL benchmark tasks \citep{liu2023datasets} and our main findings are as follows. First, even the simplest version of our approach with two arms ($K$=2) is quite effective and results in state-of-the-art performance. Second, our approach achieves excellent performance for the challenging setting of small cost constraint thresholds. Third, the performance improves with higher arms ($K$ > 2) but shows diminishing returns beyond $K$=5. Finally, O3SRL performs effectively with different offline RL algorithms and higher cost limits demonstrating its robust performance.

\noindent {\bf Contributions.} The key contribution of this paper is the development and evaluation of the O3SRL framework to solve offline safe RL problems. Specific contributions include:
\begin{itemize}
    \item Formulating OSRL as a minimax optimization problem and providing an iterative framework based on no-regret algorithms with convergence guarantees to the minimax solution.
    \item Development of a practical algorithm with convergence guarantee by avoiding the usage of off-policy evaluation (can lead to error propagation) and running offline RL algorithm to convergence (high computational cost) inside each iteration of a multi-round method.
    \item Empirical evaluation of the practical algorithm and its variants on DSRL benchmark tasks to demonstrate its effectiveness over state-of-the-art methods.
\end{itemize}

%% file: problem-setup.tex
\section{Problem Setup}

RL problems with safety constraints are typically formulated using the Constrained Markov Decision Process (CMDP) framework \citep{altman2021constrained}. A CMDP is defined by the tuple $(\mathcal{S}, \mathcal{A}, P, r, c, \gamma, \mu)$, where $\mathcal{S}$ and $\mathcal{A}$ denote the state and action spaces respectively, $P: \mathcal{S} \times \mathcal{A} \times \mathcal{S} \rightarrow [0,1]$ is the unknown stochastic state transition function, $r: \mathcal{S} \times \mathcal{A} \rightarrow \mathbb{R}$ is the reward function, $c: \mathcal{S} \times \mathcal{A} \rightarrow \mathbb{R}_{\geq 0}$ is the cost function, $\gamma \in (0,1)$ is the discount factor, and $\mu$ is the initial state distribution. 

Let $\pi: \mathcal{S} \rightarrow \mathcal{A}$ denote a policy and let $\tau = \{(s_t, a_t, r_t, c_t)\}_{t=1}^{T}$ be a trajectory of length $T$ induced by $\pi$ under dynamics $P$. The cumulative discounted return and cost for a trajectory $\tau$ are given by $R(\tau) = \sum_{t=1}^{T} \gamma^t r_t$ and $C(\tau) = \sum_{t=1}^{T} \gamma^t c_t$, respectively. In offline safe reinforcement learning (OSRL) problems, the agent is given access only to a static dataset $\mathcal{D}_{OSRL} = \{(s_i, a_i, r_i, c_i, s'_i)\}_{i=1}^{n}$ of $n$ examples collected from an unknown behavior policy, and the overall goal is to learn a policy $\pi$ to maximize the expected reward while satisfying a given constraint on the expected cumulative cost.  Formally, the learning objective is as follows:
\begin{align*}
\max_{\pi} \mathbb{E}_{\tau \sim \pi}[R(\tau)] \quad \mathrm{subject\ to} \quad \mathbb{E}_{\tau \sim \pi}[C(\tau)] \leq \kappa,
\end{align*}
where $\kappa \geq 0$ is the maximum cost threshold/limit for safety constraint. This problem setting combines the challenges of offline RL (e.g., out-of-distribution states) with the added requirement of ensuring cost constraint satisfaction from static data alone, without online interaction. 

Most of the prior work in OSRL cannot handle the setting where the cost threshold $\kappa$ is small \citep{zheng2024safe} that is important for safety-critical applications with a tight budget. 
One of the key goals of this paper is to handle this challenging setting in a principled manner while solving the general OSRL problem effectively for higher cost constraint thresholds. 

%% file: related-work.tex
\section{Related Work}

\noindent {\bf Offline RL.} This setting differs from the standard online RL setup in the following way: the goal is to learn reward maximizing policies from a given static dataset without interacting with the environment. The key technical challenge is to effectively handle out-of-distribution (OOD) states and actions  \citep{levine2020offline, offline_rl_survey_2}. Prior work has addressed this OOD challenge using improvements in value function estimation \citep{fujimoto2021minimalist, kostrikov2022offline, kumar2019stabilizing, lyu2022mildly, yang2022rorl,nguyen2023viper,nguyen-tang2023on}, sequential modeling \citep{janner2021offline, wang2022bootstrapped}, uncertainty-awareness \citep{an2021uncertainty, bai2022pessimistic}, constraining policy based on divergence \citep{wu2019behavior, jaques2019way, wu2022supported}, by extending methods for model based RL \citep{kidambi2020morel, yu2020mopo, rigter2022rambo} and imitation learning \citep{xu2022policy}, and action selection based on filtering and reweighting \citep{chen2023offline, hansen2023idql}.

\noindent {\bf Offline safe RL.} This is a generalization of the offline RL setting where we need to learn a safe policy from a given offline dataset and a pre-specified cost constraint \citep{liu2023datasets}. Offline safe RL is a relatively less-studied problem. Prior work includes methods based on constrained policy optimization \citep{polosky2022constrained, lee2022coptidice} and Lagrangian relaxation \citep{le2019batch, xu2022constraints}. \citep{le2019batch} proposed one of the first batch-constrained frameworks that combines offline RL with online Lagrange-multiplier updates. However, this approach faces practical challenges (error propagation and high computational effort) due to the need to call an OPE procedure multiple times in each iteration which we address in this paper. We elaborate on these challenges and differences in the Appendix. 

Extensions with convex MDP assumptions have also been explored \citep{zhang2024safe}, along with approaches based on adversarial cost penalties \citep{wei2024adversarially}, decision transformer \citep{liu2023constrained}, and diffusion models to create safe policies \citep{lin2023safe,  zheng2024safe, yao2024oasis}.  CAPS \citep{chemingui2025constraint} addresses dynamic safety constraints at deployment time by switching between a set of pre-trained policies based on the test-time cost budget. 
\citep{guo2025constraint} introduced CCAC framework, which conditions the policy and value functions on cost thresholds to enable adaptation and generalization to unseen constraint levels. 
FISOR \citep{zheng2024safe} addresses safety by enforcing hard constraints—ensuring state-wise cost constraint satisfaction. Rather than optimizing toward a given cost limit, FISOR focuses on minimizing cost directly by learning policies that select actions only within the feasible region.

This paper aims to address the following two drawbacks of prior work: 
{\bf 1)} Lagrangian based constrained policy optimization methods solve inter-twined optimization problems and are typically unstable in practice resulting in poor performance; and  {\bf 2)}  While  FISOR produces safe policies under tight cost constraints, it achieves low reward performance.
The proposed O3SRL framework provides a theoretically-grounded and yet practically-effective solution that can be wrapped around existing offline RL methods and does not require an OPE procedure.  

%% file: technical-approach.tex
\section{O3SRL Framework for Offline Constrained RL}

In this section, we first provide a minimax optimization formulation of the offline constrained RL problem. Next, we describe the general O3SRL framework based on no-regret algorithms to solve this optimization problem along with theoretical guarantees for convergence to the minimax solution. Finally, we provide an approximate algorithm for practical purposes.

\subsection{Minimax Optimization Formulation}

The discounted value functions of a policy $\pi$ for reward and cost objectives are defined as follows.
\begin{align*}
    V^{\pi}_r := \mathbb{E}_{\pi} \left[\sum_{t=0}^{\infty} \gamma^t r(s_t,a_t) \right] \quad \mathrm{and} \;\; V^{\pi}_c := \mathbb{E}_{\pi} \left[\sum_{t=0}^{\infty} \gamma^t c(s_t,a_t) \right]
\end{align*}
where $\gamma$ is the discount factor, $r(s_t,a_t)$=$r_t$ and $c(s_t,a_t)$=$c_t$ are the reward and cost values corresponding to state-action pair $(s_t,a_t)$ respectively.

Let $\Pi$ be the set of policies and $\Delta \Pi$ be the set of all distributions over $\Pi$. We aim to solve the following constrained optimization problem: 
\begin{align}
    \max_{D \in \Delta \Pi} \mathbb{E}_{\pi \sim D} \left[ V^{\pi}_r \right] \text{ subject to } \mathbb{E}_{\pi \sim D} \left[ V^{\pi}_c \right] \leq \kappa. 
    \label{eq: primal problem}
\end{align}

Let $L(D, \lambda)$ be the Lagrangian of \Cref{eq: primal problem} where $\lambda$ is the Lagrange multiplier.
 \begin{align}
     L(D, \lambda) := \mathbb{E}_{\pi \sim D} \left[ V^{\pi}_r \right] - \lambda (\mathbb{E}_{\pi \sim D} \left[ V^{\pi}_c \right]  - \kappa) = \mathbb{E}_{\pi \sim D} \left[ V^{\pi}_{r - \lambda(c - (1 - \gamma) \kappa)} \right]. 
     \label{Lagrangian}
 \end{align}

 The dual problem of \Cref{eq: primal problem} is given by: 
 \begin{align}
     \min_{\lambda \geq 0} \max_{D \in \Delta \Pi}  L(D,\lambda). 
     \label{eq: dual}
 \end{align}
 Note that the objective and the constraint in \Cref{eq: primal problem} are linear, thus convex and differentiable. Additionally, the constraint in \Cref{eq: primal problem} satisfies Slater's condition when $\Pi$
 is expressive enough that there exists at least one policy 
 $\pi \in \Pi$ that is strictly safe ($V^{\pi}_c \leq \kappa$), thus strong duality holds. This implies that solving the primal problem in \Cref{eq: primal problem} is equivalent to solving the dual problem in \Cref{eq: dual}. It is also equivalent to constraining the domain of $\lambda$ to any bounded domain $\Lambda$ = $[0,C]$ for some $C > 0$: if $(D^*, \lambda^*)$ is a solution to the dual problem and if $\lambda^* > 0$, then $(D^*, 1)$ is also a solution to the dual problem. This is because under strong duality, any solution $(D^*, \lambda^*)$ of the dual must satisfies the KKT conditions:  $\lambda (\mathbb{E}_{\pi \sim D} \left[ V^{\pi}_c \right]  - \kappa) = 0$. 

\begin{defn}[$\epsilon$-approximate equilibrium]
$(\hat{D}, \hat{\lambda}) \in \Delta \Pi \times \Lambda$ is $\epsilon$-approximate minimax equilibrium of \Cref{eq: dual} if 
    \begin{align*}
        L(D, \hat{\lambda}) - \epsilon \leq L(\hat{D}, \hat{\lambda}) \leq L(\hat{D}, \lambda) + \epsilon, \forall (D,\lambda) \in \Delta \Pi \times \Lambda.  
    \end{align*}
    \label{eq: approximate equilibrium}
\end{defn}

\subsection{General O3SRL Framework via No-Regret Algorithms}

\noindent {\bf Overview of the O3SRL Framework.} O3SRL relies on two key elements: 1) an offline RL oracle which takes a training dataset of (state, action, next-state, and reward) tuples to produce a policy that maximizes reward, and 2) a no-regret algorithm which adaptively selects the value of Lagrange variable $\lambda$ from a given search space $\Lambda$. We initialize $\lambda$ with $\lambda_0$ and the distribution over policies $D$ to $D_0$ respectively. In each iteration $t$, we perform two algorithmic steps. First, we employ the given offline safe RL training dataset $\mathcal{D}_{OSRL} = \{(s_i, a_i, r_i, c_i, s'_i)\}_{i=1}^{n}$ of $n$ examples and convert it into a corresponding offline RL training dataset $\mathcal{D}_{ORL}^{t} = \{(s_i, a_i, r'_i, s'_i)\}_{i=1}^{n}$ by defining a new reward $r'$ that combines original reward $r$ and cost $c$ values using $\lambda_t$ based on the Equation \ref{Lagrangian}: $r'_i$ = $r_i - \lambda_{t-1}(c_i - (1- \gamma) \kappa)$ where $\gamma$ is the discount factor and $\kappa$ is the cost constraint threshold. We call an offline RL oracle on the dataset $\mathcal{D}_{ORL}^{t}$ to get the distribution over policies $D_t$. Second, we call a no-regret algorithm \citep{OCO} by passing $\lambda_{t-1}$ and $D_t$ to get the updated Lagrange variable $\lambda_{t}$. At the end of $T$ iterations, we return the average distribution of policies $\bar{D}$ and the average Lagrangian value $\bar{\lambda}$ as the solution. Algorithm \ref{algorithm: osrl via no-regret} provides a pseudo-code of the O3SRL framework.

 \begin{algorithm}[h!]
     \begin{algorithmic}[1]
         \Require Offline training data $\mathcal{D}_{OSRL} = \{(s_i, a_i, r_i, c_i, s'_i)\}_{i=1}^{n}$, Offline RL oracle $\OfflineOracle$, no-regret update oracle $\noregret$, and search space of Lagrange variable $\Lambda$
         \State Initialize $\lambda_0 \in \Lambda, D_0 \in \Delta \Pi$ 
         \For{$t = 1, \ldots, T$}
            \State $D_t \leftarrow \OfflineOracle \left( \mathcal{D}_{ORL}^t = \{(s_i, a_i, r'_i = r_i - \lambda_{t-1}(c_i - (1- \gamma) \kappa , s'_i)\}_{i=1}^{n} \right)$
            \State $\lambda_{t} \leftarrow \noregret(\lambda_{t-1}, D_t, \Lambda)$
            \label{line: return of offlinerl oracle}
         \EndFor
         \Ensure $\bar{D} = \frac{1}{T} \sum_{t=1}^T D_t$ and $\bar{\lambda} = \frac{1}{T} \sum_{t=1}^T \lambda_t$  \Comment{Parameter averaging} 
     \end{algorithmic}
     \caption{General O3SRL Framework for Offline Constrained RL via No-Regret Algorithms}
          \label{algorithm: osrl via no-regret}
 \end{algorithm}

In what follows, we first formally define offline RL oracle and no-regret algorithm. Next, we provide a theoretical result that guarantees that the O3SRL framework converges to the minimax solution.

\vspace{0.5ex}

 \begin{defn}[Offline RL oracle]
 Let $\OfflineOracle$ be an offline RL oracle that takes the offline data $\mathcal{D}^n_{f}$ of size $n$ from an MDP with the expected reward functions $f$ (and state transition function $P$) and produces a policy distribution $D = \OfflineOracle(\mathcal{D}^n_f) \in \Delta \Pi$ at the estimation error $\epsoffline(n)$, i.e., 

 \begin{align}
\mathbb{E}_{\mathcal{D}^n_f}\mathbb{E}_{\pi \sim \OfflineOracle(\mathcal{D}^n_f)} \left[ V^{*}_{f} - V^{\pi}_f \right] \leq \epsoffline(n). 
\label{eq: offline rl error}
 \end{align}
 \label{defn: offlinerl oracle}
 \end{defn}

\noindent {\bf Remark 1.} The estimation error of an offline RL algorithm $\epsoffline(n)$ also depends on the coverage of the offline data for the optimal policy. This coverage in turn depends on the underlying MDP class, i.e., the stochastic transition function and reward function. In \Cref{algorithm: osrl via no-regret}, we employ an offline RL algorithm repeatedly for multiple MDPs with different reward functions. However, the same order rate $\epsoffline(n)$ can be applied to multiple such offline RL guarantees, with perhaps at the cost of a constant factor. The main reason is that, most notions of coverage depend very mildly on the reward functions, if at all. For example, the concentrability coefficients \citep{liu2020off,rashidinejad2021bridging} do not depend on the reward function at all.  Similarly, the data diversity measure in \citep{nguyen-tang2023on} and the policy transfer coefficient \citep{nguyen2024statistical} do not depend on the specific reward function but rather a class of reward functions. In particular, if we use $\mathcal{F}$ to approximate $V_r^{\pi}$ and $\mathcal{G}$ to approximate $V_c^{\pi}$, then $\mathcal{F} - \Lambda \cdot \mathcal{G}: = \{f - \lambda \cdot g: f \in \mathcal{F}, g \in \mathcal{G}, \lambda \in \Lambda\}$ suffices to control the \underline{same} policy transfer coefficient of a class of MDPs defined by any newly created reward functions used in \Cref{algorithm: osrl via no-regret}. Additionally, under good data coverage, we typically have $\epsoffline(n) = \mathcal{O}(\frac{1}{\sqrt{n}})$.

 \begin{defn}[No-regret algorithm]
     Let $\noregret$ is a no-regret algorithm for $\lambda \in \Lambda$ with regret bound $R_T(\Lambda)$, i.e., 
     \begin{align*}
         \lambda_{t+1} = \noregret(\lambda_t, D_t, \Lambda),
     \end{align*}
     where
     \begin{align}
         \sum_{t=1}^T L(D_t, \lambda_t) - \min_{\lambda \in \Lambda}\sum_{t=1}^T L(D_t, \lambda) \leq R_T(\Lambda)
         \label{eq: no-regret error}
     \end{align}
     and $\lim_{T \rightarrow \infty} \frac{R_T(\Lambda)}{T} \rightarrow 0$. 
     \label{defn: no-regret oracle}
 \end{defn}

We consider a generic framework for solving offline constrained RL as shown in Algorithm \ref{algorithm: osrl via no-regret}. It relies on an offline RL oracle as per Definition \ref{defn: offlinerl oracle} and a no-regret update oracle as per Definition \ref{defn: no-regret oracle}. 

 \noindent {\bf Remark 2.} From the given offline safe RL training dataset $\mathcal{D}_{OSRL} = \{(s_i, a_i, r_i, c_i, s'_i)\}_{i=1}^{n}$ of $n$ examples, we can create a corresponding dataset that simulates the experiences of any MDP with the same unknown state transition probability function and an arbitrary reward function $r' = f(r,c)$, by simply computing $r'_i = f(r_i,c_i)$. Specifically, in our algorithm we define the modified reward function based on the Equation \ref{Lagrangian} as: $r'$ = $r - \lambda_t(c - (1- \gamma) \kappa)$ where $\lambda_t \in \Lambda$ is the Lagrange variable, $\gamma$ is the discount factor, and $\kappa$ is the cost constraint threshold.

The below theorem provides guarantee that \Cref{algorithm: osrl via no-regret} will converge to the minimax solution. 

\vspace{0.5ex}

 \begin{theorem}
     Let $(\bar{D}, \bar{\lambda})$ be the output of \Cref{algorithm: osrl via no-regret}. Then, $(\bar{D}, \bar{\lambda})$ is $\epsilon$-approximate equilibrium, where $\epsilon = \epsoffline(n) + \frac{R_T(\Lambda)}{T}$.
     \label{theorem: main}
 \end{theorem}
See the detailed proof in \Cref{section: proof}.

\section{Practical Algorithms}
\label{sec:practice}

In this section, we introduce an approximation to Algorithm~\ref{algorithm: osrl via no-regret} that addresses two key practical challenges.  Below we outline these challenges and the corresponding modifications. The resulting Algorithm \ref{algorithm:approx} with minor approximation forms the basis of our primary empirical contribution.

The first practical challenge is that Algorithm~\ref{algorithm: osrl via no-regret} assumes access to an offline RL oracle, which is computationally expensive to approximate by running offline RL algorithms to convergence. This is especially problematic because the oracle must be invoked in every round of the algorithm. To mitigate this cost, we assume a \underline{weaker} oracle: a \emph{stochastic oracle} for offline RL in Assumption~\ref{assumption: stochastic oracle}. This assumption says that we have access to an oracle that returns a near-optimal policy and a stochastic approximation of the value {function} of this near-optimal policy. 

\begin{defn}
    Let $\StochasticOfflineOracle$ be a stochastic oracle for offline RL that takes the offline data $\mathcal{D}^n_f$ of size $n$ from an MDP with the expected reward function $f$ (and state transition function $P$ across all MDP instances we have considered so far) and produces a policy distribution and a stochastic approximation of the value of the optimal policy: $(\widetilde{D}, \widetilde{V}) = \StochasticOfflineOracle(\mathcal{D}^n_f)$, where 
    \begin{align}
    \mathbb{E}_{\mathcal{D}^n_f}\mathbb{E}_{\pi \sim \widetilde{D}} \left[ V^{*}_{f} - V^{\pi}_f \right] \leq \epsoffline(n), \text{ and } \mathbb{E}[\widetilde{V}] = \mathbb{E}_{\pi \sim \tilde{D}} \left[ V_f^{\pi} \right]. 
    \label{eq: offline rl error for stochastic oracle}
     \end{align}
     \label{assumption: stochastic oracle}
\end{defn}

This significantly relaxes the need to return an exact value of the value function of a near-optimal policy by an offline RL oracle, and, more importantly, leaves room for further practical approximation (as we elaborate in \Cref{subsection: more practical}).

The second practical challenge arises from the use of no-regret algorithms over a continuous range of $\lambda$. {The online optimization over a continuous domain of $\lambda$ requires an off-policy estimate of a target policy under a mixed reward of the form $r - \lambda c$ for \emph{every} value of $\lambda$. This often requires an OPE procedure}
(to compute $V^{D_t}_{c - (1-\gamma) \kappa}$ where $D^t$ is the near-optimal policy distribution returned at line~\ref{line: return of offlinerl oracle} in \Cref{algorithm: osrl via no-regret}) at every iteration. {The application of OPE at every iteration is problematic in two ways: 1) Can compound the OPE estimation error exponentially over the length of the iterations, causing instability and large error; and 2) Employing OPE for every iteration is computationally expensive.}
 
To address this, we discretize the range of $\lambda$ into $K$ discrete values $\{\lambda^{(1)},\ldots, \lambda^{(K)}\}$ and optimize over this finite set. This transforms the problem into a multi-armed bandit (MAB) setting, where each of the $K$ arms corresponds to a particular $\lambda$ value. Crucially, many MAB algorithms do not require estimates of the underlying value functions, thus avoiding the need for OPE. In this work, we adopt the EXP3 algorithm~\citep{Auer02}, which maintains a distribution over arms that is updated and sampled from in each round of Algorithm~\ref{algorithm:approx}.

\begin{algorithm}
    \begin{algorithmic}[1]
        \Require Stochastic oracle for offline RL $\StochasticOfflineOracle$, Offline safe RL training data $\mathcal{D}_{OSRL} = \{(s_i, a_i, r_i, c_i, s'_i)\}_{i=1}^{n}$, and search space $\Lambda = \{\lambda^{(1)}, \ldots, \lambda^{(K)}\}$
        \State Initialize $P_0(\lambda) = \frac{1}{K}$, $\bar{c}=\kappa/H$, where $\kappa$ is cost limit and $H=\frac{1}{1-\gamma}$ is the horizon
        \For{$t = 1,\ldots, T$}
        \State $(\tilde{D}_t, \tilde{V}_t) = \StochasticOfflineOracle \left(\mathcal{D}_{ORL}^t = \{(s_i, a_i, r'_i = r_i - \lambda_{t-1}(c_i - (1- \gamma) \kappa , s'_i)\}_{i=1}^{n} \right)$
        \State Sample $\lambda_{t} \sim P_{t}$ where $P_{t}(\lambda) \propto P_{t-1}(\lambda) \exp \left( -\eta  \frac{\tilde{V}_t 1\{\lambda=\lambda_{t-1}\}}{P_{t-1}(\lambda)}\right)$
        \EndFor 

        \Ensure $\bar{D} = \text{Proj}_{\Delta \Pi} \left(\frac{1}{T} \sum_{t=1}^T \widetilde{D}_t \right)$, $\bar{\lambda} = \text{Proj}_{\Lambda} \left(\frac{1}{T} \sum_{t=1}^T \lambda_t \right)$
    \end{algorithmic}
    \caption{O3SRL Approximation via Stochastic Oracle and EXP3 Multi-Arm Bandit Strategy}
     \label{algorithm:approx}
\end{algorithm}

\begin{theorem}
    Let $\Lambda = \{\lambda^{(1)}, \ldots, \lambda^{(K)}\}$, where $\lambda^1 = 0$ and $\lambda^{(i+1)} - \lambda^{(i)} = \frac{C}{K}$. Then, the solution $(\bar{D}, \bar{\lambda})$ returned by Algorithm~\ref{algorithm:approx} is $\epsilon$-approximate minimax equilibrium of \Cref{eq: dual}, where 
    \begin{align*}
     {\small
        \epsilon = \mathcal{O}\bigg( \underbrace{\epsoffline(n)}_{\text{error rate of offline RL}} + \underbrace{\sqrt{\frac{K}{T}}}_{\text{regret of EXP3}} + \underbrace{\frac{1}{K}}_{\text{discretization error}} \bigg). 
        }
    \end{align*}
    \label{theorem: error rate for exp3}
\end{theorem}

\noindent {\bf Remark 3:} If we set $K = \Theta(T^{1/3})$ and assume the minimax rate $\frac{1}{\sqrt{n}}$ of offline RL algorithm, then the error rate in \Cref{theorem: error rate for exp3} will be in the order of  $\frac{1}{\sqrt{n}} + \frac{1}{\sqrt[3]{T}}$. This rate is a bit slower than the rate $\frac{1}{\sqrt{n}} + \frac{1}{\sqrt{T}}$ had we have applied a no-regret algorithm such as Follow-The-Regularized-Leader \citep{OCO} over continuous values of $\lambda$. 

\subsection{Final Approximate Algorithm for Empirical Evaluation}
\label{subsection: more practical}

We consider an even more practical version of \Cref{algorithm:approx} to form the final algorithm for our empirical evaluation. Specifically, we make two changes.

\begin{itemize}
\item We further approximate the stochastic oracle by leveraging the fact that most offline RL algorithms are based on stochastic gradient descent. In our approximation (Algorithm \ref{algorithm:approx}), rather than running the offline RL algorithm to convergence at round $t$, we perform only $M$ gradient updates, after initializing from the result at the end of round $t-1$. We find this truncated update scheme to be effective even for small values of $M$. 

\item \Cref{algorithm:approx} (and Algorithm \ref{algorithm: osrl via no-regret}) maintains an averaged distribution over policies across iterations that is returned at the end of $T$ rounds. In our case, this would involve storing the policy produced on each round, which would be prohibitively memory intensive. Instead, we simply return the last-iterate policy $\pi_T$. 

\end{itemize}

In our experiments, we have found that this set of approximations leads to state-of-the-art performance even for as few as $K$=2 arms and a small number of Offline RL iterations $M$ per round.

%% file: experiments.tex
\section{Experiments and Results}

In this section, we experimentally evaluate the performance of the approximate O3SRL algorithm described in Section \ref{subsection: more practical}, compare with state-of-the-art baselines, and discuss the results. 

\subsection{Experimental Research Questions}

Our experiments are designed to answer the following key questions:

\vspace{0.75ex}

\textbf{Q1}: What is the influence of the number of arms in our EXP-3 based approximate O3SRL approach on both reward and safety objectives?

\vspace{0.75ex}

\textbf{Q2}: How effectively does  \ourmethod  satisfy safety/cost constraints and achieve high reward performance for small cost constraint thresholds?

\vspace{0.75ex}

\textbf{Q3}: How does the safety and reward trade-off of \ourmethod vary as the cost threshold increases?
\vspace{0.75ex}

\textbf{Q4}: Is \ourmethod compatible with different offline RL algorithms, demonstrating its generality?

\subsection{Experimental Setup}

\textbf{Benchmarks.} We evaluate \ourmethod and baseline methods using the DSRL Bullet benchmark \citep{liu2023datasets}, which provides standardized offline datasets for safe RL research and evaluation. We consider eight continuous control tasks, comprising four ``Run'' and four ``Circle'' environments. Each task involves a different agent interacting with either the Run or Circle objective, following the common Agent-Task naming convention (e.g., Car-Run, Drone-Circle).

In the Run tasks, agents are rewarded for moving quickly between two boundaries while adhering to safety constraints, such as velocity limits and positional bounds. In the Circle tasks, agents are encouraged to follow a circular trajectory within a restricted safe region. These are challenging environments which require agents to balance performance with constraint satisfaction.

\textbf{Evaluation metrics.}
Following the DSRL evaluation protocol \citep{liu2023datasets}, we evaluate all methods using normalized cumulative reward and normalized cumulative cost.   Let \( r_{\max}(M) \) and \( r_{\min}(M) \) denote the maximum and minimum empirical reward returns achievable for task \( M \). 
The normalized reward is computed as \( R_{\text{normalized}} = \frac{R_\pi - r_{\min}(M)}{r_{\max}(M) - r_{\min}(M)} \), where \( R_\pi \) is the cumulative reward of policy \( \pi \). The normalized cost is given by \( C_{\text{normalized}} = \frac{C_\pi}{\kappa} \), with \( \kappa \) being the cost budget.

where \( C_\pi \) is the cumulative cost of the policy $\pi$ and \( \kappa \) is the cost threshold. A policy $\pi$ is considered safe if \( C_{\text{normalized}} \leq 1 \). Our O3SRL method is general-purpose and can be trained for any given cost threshold $\kappa$. However, we report our main results under a stringent low cost threshold of $\kappa$=5 to highlight the method’s effectiveness in highly constrained settings. We report both mean and standard deviations over three random seeds, with each seed's policy evaluated across twenty episodes.

\textbf{Baseline methods.}
We compare O3SRL with several state-of-the-art offline safe RL methods. {\bf (1)} BC-Safe is a behavior cloning baseline trained exclusively on safe trajectories that satisfy the given cost limit. {\bf (2)} BEAR-Lag is a Lagrangian-based extension of BEAR \citep{kumar2019stabilizing}, which incorporates safety constraints during training. {\bf (3)} CPQ \citep{xu2022constraints} penalizes unsafe actions by treating out-of-distribution actions as inherently risky and updates its Q-function using only safe transitions. {\bf (4)} COptiDICE \citep{lee2022coptidice} builds on OptiDICE \citep{lee2021optidice} by applying stationary distribution correction under cost constraints.  {\bf (5)} CDT \citep{liu2023constrained} is a transformer-based method that conditions policies on returns and costs, allowing constraint-aware sequence modeling. 
{\bf (6)} CCAC \citep{guo2025constraint} learns a policy that adapts to changing safety budgets by jointly modeling constraints and environment dynamics. This approach conditions both the policy and value functions on constraint information.
{\bf (7)} CAPS \citep{chemingui2025constraint}, similar to CDT and CCAC, supports test-time adaptation to varying cost constraints by switching between a set of pre-trained policies with different cost and reward trade-offs.
{\bf (8)} FISOR \citep{zheng2024safe} is a diffusion-based method that explicitly optimizes for feasibility, aims to produce zero-violation policies, and is the state-of-the-art method for stringent cost constraints. 
 
All baselines are evaluated under the same cost constraint threshold $\kappa$ = 5. All results are averaged over three random seeds, with each policy evaluated across 20 episodes. Additional details are in the Appendix.

\textbf{Configuration of \ourmethod.}
We implement the approximate O3SRL algorithm described in Section \ref{subsection: more practical} using TD3+BC \citep{fujimoto2021minimalist} as the underlying offline RL algorithm for our main results. 
We set the search space of Lagrange variable $\Lambda$= [0, $C$=5] and use $K$=5 arms for the main results but we also present ablation results for $K$=2, 5, and 10. We observe that increasing $K$ beyond 5 (say to 10) yields only marginal improvements compared to the gains observed when increasing from $K$=2 to $K$=5.
The algorithm is trained for $T=100{,}000$ iterations, with arm probabilities updated every $M=10$ steps of the base offline RL algorithm (i.e., $M$=10 stochastic gradient updates of the policy). 
For baselines, we employ the DSRL \citep{liu2023datasets} implementation wherever available. 
For CCAC, CAPS and FISOR, we use the publicly released code provided by their respective authors. We employ IQL based instantiation of CAPS in our evaluation. Full architectural and hyperparameter details are provided in the Appendix. 

\subsection{Results and Discussion}

\noindent {\bf Performance of \ourmethod vs. the number of arms $K$.} To evaluate the impact of the granularity of discretization of \(\lambda\), we test \ourmethod using \( K \in \{2, 5, 10\} \) arms. Table \ref{tab:num_arms_results} shows the corresponding results. Across all six benchmark tasks, \ourmethod satisfies the cost constraint under every configuration and maintains consistently high reward, demonstrating that the EXP3 strategy is effective even with a coarse discretization of \(\lambda\). With just two arms, the method keeps average cost at or near zero, but tends to underperform slightly in terms of rewards. Increasing to ten arms improves rewards marginally. The five-arm configuration achieves a good trade-off: it either matches or exceeds the highest feasible reward across all tasks while keeping costs safely below the threshold. For this reason, we adopted \( K = 5 \) configuration of the approximate \ourmethod approach for all main results.

\begin{table*}[ht!]
\centering
\caption{\ourmethod results for different values of $K$: the number of arms. Each value shows the average reward ($\uparrow$) and cost ($\downarrow$). Higher reward is better, and lower cost (up to threshold 1) is better.}
\label{tab:num_arms_results}
\begin{adjustbox}{max width=\textwidth}
\begin{tabular}{lcccccccccccccc}
\toprule
\textbf{\# Arms} 
& \multicolumn{2}{c}{\textbf{BallRun}} 
& \multicolumn{2}{c}{\textbf{CarRun}} 
& \multicolumn{2}{c}{\textbf{AntRun}} 
& \multicolumn{2}{c}{\textbf{BallCircle}} 
& \multicolumn{2}{c}{\textbf{CarCircle}} 
& \multicolumn{2}{c}{\textbf{AntCircle}} \\
\cmidrule(lr){2-3} \cmidrule(lr){4-5} \cmidrule(lr){6-7} \cmidrule(lr){8-9} 
\cmidrule(lr){10-11} \cmidrule(lr){12-13}  
& reward $\uparrow$ & cost $\downarrow$ 
& reward $\uparrow$ & cost $\downarrow$ 
& reward $\uparrow$ & cost $\downarrow$ 
& reward $\uparrow$ & cost $\downarrow$ 
& reward $\uparrow$ & cost $\downarrow$ 
& reward $\uparrow$ & cost $\downarrow$ \\
\midrule

$K=2$ & \textbf{0.20} & \textbf{0.00} & \textbf{0.96} & \textbf{0.01} & \textbf{0.24} & \textbf{0.03} & \textbf{0.62} & \textbf{0.00} & \textbf{0.64} & \textbf{0.00} & \textbf{0.44} & \textbf{0.04} \\

$K=5$ & \textbf{0.25} & \textbf{0.00} & \textbf{0.96} & \textbf{0.02} & \textbf{0.33} & \textbf{0.14} & \textbf{0.62} & \textbf{0.06} & \textbf{0.66} & \textbf{0.11} & \textbf{0.48} & \textbf{0.00} \\
$K=10$ & \textbf{0.27} & \textbf{0.00} & \textbf{0.96} & \textbf{0.00} & \textbf{0.29} & \textbf{0.17} & \textbf{0.64} & \textbf{0.28} & \textbf{0.67} & \textbf{0.15} & \textbf{0.49} & \textbf{0.00} \\

\bottomrule
\end{tabular}
\end{adjustbox}
\end{table*}

\begin{table}[htbp]
\centering
\caption{\ourmethod normalized rewards and costs results. The cost threshold is 1. The $\uparrow$ symbol denotes that higher rewards are better. The $\downarrow$ symbol denotes that lower costs (up to threshold 1) are better. Each value is averaged over 20 evaluation episodes, and three random seeds. \textbf{Bold}: Safe agents whose normalized cost $\leq$ 1. \textcolor{gray}{Gray}: Unsafe agents. \textbf{\textcolor{blue}{Blue}}: Safe agents with the highest reward.}
\small
\begin{adjustbox}{width=\textwidth}
\begin{tabular}{ll|cccccccccc}
\toprule
\textbf{Tasks} &  & BC Safe & BEAR-Lag & CPQ & COptiDICE & CDT & CCAC & CAPS & FISOR & O3SRL \\
\midrule
\multirow{2}{*}{BallRun} & Reward $\uparrow$ & \textcolor{gray}{\meanstd{0.16}{0.11}} & \textcolor{gray}{\meanstd{0.53}{0.47}} & \textcolor{gray}{\meanstd{0.09}{0.26}} & \textcolor{gray}{\meanstd{0.53}{0.10}} & \textcolor{gray}{\meanstd{0.27}{0.09}} & \textcolor{blue}{\textbf{\meanstd{0.31}{0.01}}} & \textbf{\meanstd{0.07}{0.05}} & \textcolor{gray}{\meanstd{0.09}{0.07}} & {\textbf{\meanstd{0.25}{0.03}}}\\

& Cost $\downarrow$ & \textcolor{gray}{\meanstd{4.50}{3.38}} & \textcolor{gray}{\meanstd{18.60}{0.35}} & \textcolor{gray}{\meanstd{2.20}{2.88}} & \textcolor{gray}{\meanstd{10.83}{1.68}} & \textcolor{gray}{\meanstd{2.57}{3.23}} & \textcolor{blue}{\textbf{\meanstd{0.00}{0.00}}} & \textbf{\meanstd{0.00}{0.00}} & \textcolor{gray}{\meanstd{1.28}{1.70}} & {\textbf{\meanstd{0.00}{0.00}}}\\
\midrule

\multirow{2}{*}{CarRun} & Reward $\uparrow$ & \textbf{\meanstd{0.92}{0.01}} & \textcolor{gray}{\meanstd{0.25}{1.23}} & \textbf{\meanstd{0.93}{0.01}} & \textbf{\meanstd{0.91}{0.04}} & \textcolor{blue}{\textbf{\meanstd{0.99}{0.00}}} & \textcolor{gray}{\meanstd{1.82}{0.84}} & \textbf{\meanstd{0.97}{0.00}} & \textbf{\meanstd{0.74}{0.01}} & \textbf{\meanstd{0.96}{0.01}}\\

& Cost $\downarrow$ & \textbf{\meanstd{0.26}{0.23}} & \textcolor{gray}{\meanstd{30.17}{10.64}} & \textbf{\meanstd{0.20}{0.18}} & \textbf{\meanstd{0.00}{0.00}} & \textcolor{blue}{\textbf{\meanstd{0.90}{0.34}}} & \textcolor{gray}{\meanstd{24.57}{20.94}} & \textbf{\meanstd{0.11}{0.17}} & \textbf{\meanstd{0.00}{0.00}} & \textbf{\meanstd{0.02}{0.03}}\\
\midrule

\multirow{2}{*}{DroneRun} & Reward $\uparrow$ & \textcolor{gray}{\meanstd{0.41}{0.23}} & \textcolor{gray}{\meanstd{-0.21}{0.12}} & \textcolor{gray}{\meanstd{0.29}{0.11}} & \textcolor{gray}{\meanstd{0.68}{0.01}} & \textcolor{blue}{\textbf{\meanstd{0.58}{0.00}}} & \textcolor{gray}{\meanstd{0.50}{0.08}} & \textcolor{gray}{\meanstd{0.41}{0.06}} & \textcolor{gray}{\meanstd{0.31}{0.04}} & \textbf{\meanstd{0.32}{0.05}}\\

& Cost $\downarrow$ & \textcolor{gray}{\meanstd{1.62}{1.73}} & \textcolor{gray}{\meanstd{14.85}{8.77}} & \textcolor{gray}{\meanstd{2.35}{4.08}} & \textcolor{gray}{\meanstd{15.02}{0.10}} & \textcolor{blue}{\textbf{\meanstd{0.07}{0.07}}} & \textcolor{gray}{\meanstd{16.29}{9.35}} & \textcolor{gray}{\meanstd{5.70}{3.08}} & \textcolor{gray}{\meanstd{2.52}{1.10}} & \textbf{\meanstd{0.68}{1.18}}\\
\midrule

\multirow{2}{*}{AntRun} & Reward $\uparrow$ & \textcolor{gray}{\meanstd{0.56}{0.02}} & \textbf{\meanstd{0.02}{0.02}} & \textbf{\meanstd{0.03}{0.05}} & \textcolor{gray}{\meanstd{0.61}{0.01}} & \textcolor{gray}{\meanstd{0.70}{0.03}} & \textbf{\meanstd{0.03}{0.10}} & \textcolor{gray}{\meanstd{0.53}{0.13}} & \textcolor{blue}{\textbf{\meanstd{0.43}{0.02}}} & \textbf{\meanstd{0.33}{0.13}}\\

& Cost $\downarrow$ & \textcolor{gray}{\meanstd{1.15}{0.47}} & \textbf{\meanstd{0.00}{0.01}} & \textbf{\meanstd{0.05}{0.08}} & \textcolor{gray}{\meanstd{3.26}{1.39}} & \textcolor{gray}{\meanstd{1.66}{0.24}} & \textbf{\meanstd{0.00}{0.00}} & \textcolor{gray}{\meanstd{2.03}{2.17}} & \textcolor{blue}{\textbf{\meanstd{0.27}{0.15}}} & \textbf{\meanstd{0.14}{0.10}}\\
\midrule

\multirow{2}{*}{BallCircle} & Reward $\uparrow$ & \textcolor{gray}{\meanstd{0.45}{0.03}} & \textcolor{gray}{\meanstd{0.85}{0.05}} & \textcolor{gray}{\meanstd{0.56}{0.10}} & \textcolor{gray}{\meanstd{0.71}{0.01}} & \textcolor{gray}{\meanstd{0.61}{0.17}} & \textcolor{gray}{\meanstd{0.47}{0.38}} & \textbf{\meanstd{0.33}{0.02}} & \textbf{\meanstd{0.32}{0.05}} & \textcolor{blue}{\textbf{\meanstd{0.62}{0.01}}}\\

& Cost $\downarrow$ & \textcolor{gray}{\meanstd{1.21}{0.23}} & \textcolor{gray}{\meanstd{10.92}{2.02}} & \textcolor{gray}{\meanstd{1.11}{1.92}} & \textcolor{gray}{\meanstd{9.41}{0.69}} & \textcolor{gray}{\meanstd{2.03}{1.39}} & \textcolor{gray}{\meanstd{10.19}{17.65}} & \textbf{\meanstd{0.01}{0.02}} & \textbf{\meanstd{0.00}{0.00}} & \textcolor{blue}{\textbf{\meanstd{0.06}{0.07}}}\\
\midrule

\multirow{2}{*}{CarCircle} & Reward $\uparrow$ & \textcolor{gray}{\meanstd{0.31}{0.06}} & \textcolor{gray}{\meanstd{0.73}{0.05}} & \textcolor{blue}{\textbf{\meanstd{0.71}{0.02}}} & \textcolor{gray}{\meanstd{0.49}{0.01}} & \textcolor{gray}{\meanstd{0.71}{0.01}} & \textcolor{gray}{\meanstd{0.70}{0.01}} & \textbf{\meanstd{0.40}{0.03}} & \textbf{\meanstd{0.37}{0.02}} & \textbf{\meanstd{0.66}{0.03}}\\

& Cost $\downarrow$ & \textcolor{gray}{\meanstd{1.57}{0.07}} & \textcolor{gray}{\meanstd{7.39}{1.28}} & \textcolor{blue}{\textbf{\meanstd{0.00}{0.00}}} & \textcolor{gray}{\meanstd{10.82}{1.28}} & \textcolor{gray}{\meanstd{1.58}{0.57}} & \textcolor{gray}{\meanstd{1.61}{0.51}} & \textbf{\meanstd{0.03}{0.03}} & \textbf{\meanstd{0.00}{0.00}} & \textbf{\meanstd{0.11}{0.16}}\\
\midrule

\multirow{2}{*}{DroneCircle} & Reward $\uparrow$ & \textcolor{gray}{\meanstd{0.50}{0.01}} & \textcolor{gray}{\meanstd{0.84}{0.07}} & \textbf{\meanstd{-0.21}{0.04}} & \textcolor{gray}{\meanstd{0.26}{0.02}} & \textcolor{gray}{\meanstd{0.55}{0.01}} & \textbf{\meanstd{0.43}{0.04}} & \textbf{\meanstd{0.36}{0.02}} & \textbf{\meanstd{0.48}{0.01}} & \textcolor{blue}{\textbf{\meanstd{0.49}{0.07}}}\\

& Cost $\downarrow$ & \textcolor{gray}{\meanstd{1.22}{0.31}} & \textcolor{gray}{\meanstd{15.40}{2.62}} & \textbf{\meanstd{0.70}{0.41}} & \textcolor{gray}{\meanstd{3.68}{0.5}1} & \textcolor{gray}{\meanstd{1.14}{0.06}} & \textbf{\meanstd{0.06}{0.07}} & \textbf{\meanstd{0.00}{0.00}} & \textbf{\meanstd{0.17}{0.15}} & \textcolor{blue}{\textbf{\meanstd{0.23}{0.34}}}\\
\midrule

\multirow{2}{*}{AntCircle} & Reward $\uparrow$ & \textcolor{gray}{\meanstd{0.40}{0.02}} & \textcolor{gray}{\meanstd{0.67}{0.05}} & \textbf{\meanstd{0.00}{0.00}} & \textcolor{gray}{\meanstd{0.18}{0.02}} & \textcolor{gray}{\meanstd{0.45}{0.05}} & \textcolor{blue}{\textbf{\meanstd{0.49}{0.07}}} & \textbf{\meanstd{0.33}{0.05}} & \textbf{\meanstd{0.24}{0.02}} & {\textbf{\meanstd{0.48}{0.06}}}\\

& Cost $\downarrow$ & \textcolor{gray}{\meanstd{4.72}{0.87}} & \textcolor{gray}{\meanstd{19.00}{1.68}} & \textbf{\meanstd{0.00}{0.00}} & \textcolor{gray}{\meanstd{17.40}{0.36}} & \textcolor{gray}{\meanstd{6.59}{1.42}} & \textcolor{blue}{\textbf{\meanstd{0.02}{0.003}}} & \textbf{\meanstd{0.00}{0.00}} & \textbf{\meanstd{0.04}{0.08}} & {\textbf{\meanstd{0.00}{0.00}}}\\
\bottomrule
\end{tabular}
\end{adjustbox}

\label{tab:main_results}
\end{table}

\noindent {\bf O3SRL vs. Baselines.} Table \ref{tab:main_results} presents the comparative evaluation of \ourmethod against baselines across eight DSRL tasks. The table reports both normalized reward (higher is better) and normalized cost (lower is better, with values less than 1 indicating constraint satisfaction). \ourmethod is the {\em only} method that consistently satisfies the cost constraint across all eight tasks. This is particularly noteworthy given the stringent cost budget, which poses a significant challenge for all baseline algorithms. 

\ourmethod’s consistent satisfaction of the cost constraint stands in contrast to methods that only occasionally achieve strong reward, often by violating safety. For instance, CDT achieves high rewards on tasks such as \textit{CarRun} and \textit{DroneRun}, but exceeds the safety budget in the other six tasks—making it unreliable. Similarly, CCAC achieves the best rewards on \textit{BallRun} and \textit{AntCircle}, yet violates the cost constraint in four out of eight tasks, further underscoring the challenge of maintaining both high performance and safety across environments. CPQ, on the other hand, stays within the budget more often, but its performance is inconsistent: it achieves a competitive reward on \textit{CarCircle}, while falling short on rewards when it does remain safe, and it exceeds the cost threshold in three tasks.

While CAPS and FISOR maintain safe behavior in six out of eight tasks, their reward performance consistently trails behind \ourmethod. For example, in \textit{BallRun}, \ourmethod is one of the only three safe methods (along with CAPS and CCAC), yet it achieves a reward of $0.25 \pm 0.03$, which is over three times higher than CAPS ($0.07 \pm 0.05$). A similar pattern holds in \textit{BallCircle}, where \ourmethod maintains safety and attains $0.65 \pm 0.01$ reward—more than double that of CAPS ($0.23 \pm 0.17$) and FISOR ($0.27 \pm 0.15$), both of which are also safe in that task. Note that FISOR, which aims for zero cost policies, fails to meet the cost threshold in the \textit{DroneRun} and \textit{BallRun} environments.

Overall, the results in Table \ref{tab:main_results} demonstrate that \ourmethod strikes a uniquely favorable balance between maximizing rewards and satisfying safety constraints. It consistently ensures safety where many other methods fail, especially under a tight cost budget. Crucially, this safety does not come at a severe expense to the reward performance; \ourmethod is the best-performing method (in terms of reward) among safe agents in two out of eight tasks, and consistently ranks second in the rest. This suggests that \ourmethod's iterative, no-regret approach to optimizing the reward and safety trade-off is more robust and effective than direct Lagrangian optimization or specialized heuristic-based safety mechanisms in these challenging offline safe RL settings.

\noindent {\bf Performance of \ourmethod vs. increasing cost limits.} To assess the flexibility of \ourmethod in handling different safety constraints, we perform an ablation study across two extra cost limits beyond the default setting of $\kappa = 5$ used in the main experiments. Table \ref{tab:extra cost} shows the corresponding results. We find that \ourmethod continues to perform effectively under higher cost limits (i.e., less stringent safety constraints), adjusting its behavior to make better use of the available budget. As the constraint becomes more permissive, the learned policy shifts toward higher-reward strategies, while still respecting the specified cost limit. These results demonstrate that the \ourmethod method does not rely on tuning for a particular budget.

\begin{table}[ht!]
\centering
\caption{\ourmethod results for different values of $\kappa$: cost budget. Each value shows the average reward ($\uparrow$) and cost ($\downarrow$). Higher reward is better, and lower cost (up to threshold 1) is better.}
\label{tab:extra cost}
\begin{adjustbox}{max width=0.8\textwidth}
\begin{tabular}{lcccccccccc}
\toprule
\textbf{Cost Limit} 
& \multicolumn{2}{c}{\textbf{BallRun}} 
& \multicolumn{2}{c}{\textbf{DroneRun}} 
& \multicolumn{2}{c}{\textbf{BallCircle}} 
& \multicolumn{2}{c}{\textbf{DroneCircle}} \\
\cmidrule(lr){2-3} \cmidrule(lr){4-5} \cmidrule(lr){6-7} \cmidrule(lr){8-9} 
 
& reward $\uparrow$ & cost $\downarrow$ 
& reward $\uparrow$ & cost $\downarrow$ 
& reward $\uparrow$ & cost $\downarrow$ 
& reward $\uparrow$ & cost $\downarrow$ \\
\midrule

$\kappa = 5$ & \textbf{0.25} & \textbf{0.00} & \textbf{0.32} & \textbf{0.68} & \textbf{0.62} & \textbf{0.06} & \textbf{0.49} & \textbf{0.23}  \\

$\kappa = 20$ & \textbf{0.26} & \textbf{0.06} & \textbf{0.38} & \textbf{0.41} & \textbf{0.69} & \textbf{0.58} & \textbf{0.53} & \textbf{0.36} \\
$\kappa = 40$ & \textbf{0.53} & \textbf{0.71} & \textbf{0.38} & \textbf{0.44} & \textbf{0.76} & \textbf{0.64} & \textbf{0.65} & \textbf{0.85}\\

\bottomrule
\end{tabular}
\end{adjustbox}
\end{table}

\noindent {\bf \ourmethod with an alternative Offline RL algorithm.} To evaluate the generality of O3SRL with respect to the underlying offline RL algorithm, we conduct an ablation using Implicit Q-Learning (IQL) \citep{kostrikov2022offline} in place of TD3+BC, which is used in our main experiments, on the Circle tasks. Table \ref{tab:off_rl} shows the corresponding results. We observe that O3SRL combined with IQL achieves similar performance to the TD3+BC variant, both in satisfying safety constraints and attaining competitive reward. These results demonstrate that O3SRL is not tied to a specific offline RL method and can effectively incorporate different algorithms, highlighting its flexibility as a general framework for offline safe reinforcement learning.

\begin{table}[ht!]
\centering
\caption{\ourmethod results for different base offline RL methods. Each value shows the average reward ($\uparrow$) and cost ($\downarrow$). Higher reward is better, and lower cost (up to threshold 1) is better.}
\label{tab:off_rl}
\begin{adjustbox}{max width=0.8\textwidth}
\begin{tabular}{lcccccccccc}
\toprule
\textbf{Base} 
& \multicolumn{2}{c}{\textbf{BallCircle}} 
& \multicolumn{2}{c}{\textbf{CarCircle}} 
& \multicolumn{2}{c}{\textbf{DroneCircle}} 
& \multicolumn{2}{c}{\textbf{AntCircle}} \\
\cmidrule(lr){2-3} \cmidrule(lr){4-5} \cmidrule(lr){6-7} \cmidrule(lr){8-9} 
 
\textbf{Offline RL} & reward $\uparrow$ & cost $\downarrow$ 
& reward $\uparrow$ & cost $\downarrow$ 
& reward $\uparrow$ & cost $\downarrow$ 
& reward $\uparrow$ & cost $\downarrow$ \\
\midrule

TD3BC & \textbf{0.62} & \textbf{0.06} & \textbf{0.66} & \textbf{0.11} & \textbf{0.49} & \textbf{0.23} & \textbf{0.48} & \textbf{0.00} \\
IQL & \textbf{0.54} & \textbf{0.06} & \textbf{0.63} & \textbf{0.10} & \textbf{0.42} & \textbf{0.03} & \textbf{0.41} & \textbf{0.06} \\

\bottomrule
\end{tabular}
\end{adjustbox}
\end{table}

In Appendix, we provide extended experimental results including evaluation on more benchmark tasks; treating $\lambda$ as a fixed hyper-parameter; and ablation results for O3SRL over search space of $\lambda$ and discretization levels, and the number of stochastic gradient updates of offline RL ($M$).

%% file: summary.tex
\section{Summary and Future Work}

This paper developed and evaluated a novel online optimization framework called O3SRL to create safe and reward-maximizing decision policies from offline datasets. O3SRL solves a minimax optimization problem by employing an iterative approach that adaptively updates the policy distribution using an offline RL oracle and Lagrange variables in each iteration. To handle the practical challenges, we studied an approximate algorithm that works over discrete values of Lagrange variables (to avoid offline policy evaluation) and performs few gradient updates of an offline RL algorithm (for computational efficiency). Empirical results show that our approximate O3SRL algorithm outperforms state-of-the-art methods, especially for stringent safety/cost constraints. Future work includes applying O3SRL to real-world applications and extending it to the offline-to-online safe RL setting.

\section*{Acknowledgments}
The authors gratefully acknowledge the in part support by
USDA-NIFA funded AgAID Institute award 2021-67021-
35344. The views expressed are those of the authors and do
not reflect the official policy or position of the USDA-NIFA.

%% file: appendix.tex
\section{Proofs for Theorem 1 and Theorem 2}
\label{section: proof}
\begin{proof}[Proof of \Cref{theorem: main}]
     For any $D \in \Delta \Pi$, we have 
     \begin{align*}
         L(D, \bar{\lambda}) &= \frac{1}{T} \sum_{t=1}^T L(D, \lambda_t) &\text{ ($L(D,\lambda)$ is linear in $\lambda$)} \\ 
         &\leq \frac{1}{T} \sum_{t=1}^T \left( L(D_t, \lambda_t) + \epsoffline(n) \right) &\text{(due to \Cref{eq: offline rl error})}\\ 
         &\leq \frac{1}{T} \sum_{t=1}^T L(D_t, \bar{\lambda}) + \epsoffline(n) + \frac{R_T(\Lambda)}{T} &\text{(due to \Cref{eq: no-regret error})} \\ 
         &= L(\bar{D}, \bar{\lambda}) + \epsoffline(n) + \frac{R_T(\Lambda)}{T} &\text{ ($L(D,\lambda)$ is linear in $D$)}. 
     \end{align*}
     For any $\lambda \in \Lambda$, we have 
     \begin{align*}
         L(\bar{D}, \lambda) &= \frac{1}{T} \sum_{t=1}^T L(D_t, \lambda) \\ 
         &\geq \frac{1}{T} \sum_{t=1}^T L(D_t, \lambda_t) - \frac{R_T(\Lambda)}{T} &\text{(due to \Cref{eq: no-regret error})}\\ 
         &\geq \frac{1}{T} \sum_{t=1}^T L(\bar{D}, \lambda_t) - \epsoffline(n) - \frac{R_T(\Lambda)}{T} &\text{(due to \Cref{eq: offline rl error})} \\ 
         &= L(\bar{D}, \bar{\lambda}) - \epsoffline(n) - \frac{R_T(\Lambda)}{T}.
     \end{align*}
 \end{proof}

 \begin{proof}[Proof of \Cref{theorem: error rate for exp3}]
     The result directly follows from the assumption of the stochastic oracle in Definition~\ref{assumption: stochastic oracle}, the regret bound of EXP3 \citep{auer2002nonstochastic}, and the approximation error for the projection from $[0,C]$ into the discrete domain $\Lambda$ with resolution $1/K$ is $1/K$. 

     In particular, by \citep{auer2002nonstochastic}, we have 
     \begin{align*}
         \sum_{t=1}^T  L(\tilde{D}_t, \lambda_t) -  \min_{\lambda \in \Lambda}\sum_{t=1}^T L(\tilde{D}_t, \lambda) \leq H \sqrt{2} \sqrt{T K \ln K }.
     \end{align*}
     Applying the result in \Cref{theorem: main}, we have 
     \begin{align*}
         \max_{D \in \Delta \Pi}L(D, \hat{\lambda}) + \epsilon_1 \leq L(\bar{D}, \hat{\lambda}) \leq \min_{\lambda \in \Lambda} L(\bar{D}, \lambda) + \epsilon_1,
     \end{align*}
     where 
     \begin{align*}
         \epsilon_1 = \epsoffline(n) + H\sqrt{\frac{2 K \log K }{T}}, \text{ and } \hat{\lambda} := \frac{1}{T} \sum_{t=1}^T \lambda_t. 
     \end{align*}

     Note that the running average $\hat{\lambda}$ does not guarantee to be in $\Lambda$. Thus, with $\bar{\lambda} = \text{Proj}_{\Lambda}(\hat{\lambda})$ -- a projection of $\hat{\lambda}$ onto $\Lambda$, we have that, 
     \begin{align*}
         |L(D, \hat{\lambda}) - L(D, \bar{\lambda})| \leq \frac{H}{K}. 
     \end{align*}
     Overall, we have 
     \begin{small}
     \begin{align*}
                \max_{D \in \Delta \Pi}L(D, \bar{\lambda}) + \epsilon_1 + \frac{H}{K} \leq L(\bar{D}, \bar{\lambda}) \leq \min_{\lambda \in \Lambda} L(\bar{D}, \lambda) + \epsilon_1 + \frac{H}{K},
     \end{align*}
     \end{small}
     Thus, $(\bar{D}, \bar{\lambda})$ is $\epsilon$-approximate equilibrium, where 
     \begin{align*}
         \epsilon = \epsoffline(n) + H\sqrt{\frac{2 K \log K }{T}} + \frac{H}{K}.
     \end{align*}
 \end{proof}

\section{Extended Discussion of Related Work}
{
% \paragraph{Comparison with the Lagrangian-based algorithm for OSRL in \citep{le2019batch}.} 
Lagrangian formulation is quite natural for offline safe RL that both our work and \citep{le2019batch} share. \citep{le2019batch} and our work also share the general idea that we can use online optimization algorithms to tune the Lagrangian multiplier. The critical difference is in the design of efficient and effective algorithmic frameworks: our framework requires {\em one call to an offline RL oracle per loop} and {\em does not rely on any OPE oracle}. On the other hand, \citep{le2019batch} requires {\em two calls to an offline RL oracle and four calls to an OPE oracle per loop}: four calls to OPE oracle in each iteration can lead to error propagation and exponential error over the number of iterations posing stability challenges; and six calls to oracles of different nature in each iteration leads to huge computational expense posing scalability challenges. Our approximate O3SRL algorithmic framework overcomes these challenges in a principled manner.

}
\section{Ablation Results for O3SRL}

We provide additional ablation results for O3SRL beyond those presented in the main paper.

\subsection{Extended Evaluation of \ourmethod}

{ To further evaluate the generalization capability of \ourmethod, we extended our experiments to additional environments from the \textit{Gymnasium Safety Suite}, following the FISOR \citep{zheng2024safe} setup with a tight cost limit of 10. The results are summarized in Table~\ref{tab:extended_evals}.}

\begin{table}[htbp]
\centering
\caption{\ourmethod normalized rewards and costs results across SafetyGym environments. $\uparrow$ denotes that higher rewards are better; $\downarrow$ denotes that lower costs (up to threshold 1) are better. \textbf{Bold}: Safe agents whose normalized cost $\leq$ 1. \textcolor{gray}{Gray}: Unsafe agents. \textbf{\textcolor{blue}{Blue}}: Safe agents with the highest reward.}
\small
\begin{tabular}{ll|cccc}
\toprule
\textbf{Environment} &  & CAPS & FISOR & CCAC & O3SRL \\
\midrule
\multirow{2}{*}{PointCircle1} 
& Reward $\uparrow$ & \textcolor{blue}{\textbf{\meanstd{0.31}{0.05}}} & \textcolor{gray}{\meanstd{0.43}{0.05}} & \textcolor{gray}{\meanstd{0.57}{0.08}} & \textcolor{gray}{\meanstd{0.53}{0.02}} \\
& Cost $\downarrow$ & \textcolor{blue}{\textbf{\meanstd{0.94}{1.53}}} & \textcolor{gray}{\meanstd{14.93}{4.24}} & \textcolor{gray}{\meanstd{6.65}{3.51}} & \textcolor{gray}{\meanstd{1.31}{0.62}} \\
\midrule
\multirow{2}{*}{PointCircle2} 
& Reward $\uparrow$ & \textbf{\meanstd{0.44}{0.06}} & \textcolor{gray}{\meanstd{0.76}{0.05}} & \textcolor{gray}{\meanstd{0.03}{0.75}} & \textcolor{blue}{\textbf{\meanstd{0.52}{0.04}}} \\
& Cost $\downarrow$ & \textbf{\meanstd{0.10}{0.09}} & \textcolor{gray}{\meanstd{18.02}{4.17}} & \textcolor{gray}{\meanstd{3.99}{4.30}} & \textcolor{blue}{\textbf{\meanstd{0.55}{0.41}}} \\
\midrule
\multirow{2}{*}{AntVelocity} 
& Reward $\uparrow$ & \textcolor{blue}{\textbf{\meanstd{0.87}{0.04}}} & \textbf{\meanstd{0.89}{0.01}} & \textbf{\meanstd{-1.01}{0.00}} & {\textbf{\meanstd{0.45}{0.43}}} \\
& Cost $\downarrow$ & \textcolor{blue}{\textbf{\meanstd{0.19}{0.06}}} & \textbf{\meanstd{0.00}{0.00}} & \textbf{\meanstd{0.00}{0.00}} & {\textbf{\meanstd{0.05}{0.06}}} \\
\midrule
\multirow{2}{*}{HalfCheetah} 
& Reward $\uparrow$ & \textbf{\meanstd{0.86}{0.01}} & \textbf{\meanstd{0.89}{0.01}} & \textbf{\meanstd{0.91}{0.04}} & \textcolor{blue}{\textbf{\meanstd{0.93}{0.02}}} \\
& Cost $\downarrow$ & \textbf{\meanstd{0.27}{0.07}} & \textbf{\meanstd{0.00}{0.00}} & \textbf{\meanstd{0.97}{0.12}} & \textcolor{blue}{\textbf{\meanstd{0.00}{0.00}}} \\
\midrule
\multirow{2}{*}{HopperVelocity} 
& Reward $\uparrow$ & \textcolor{gray}{\meanstd{0.29}{0.14}} & \textbf{\meanstd{0.12}{0.03}} & \textbf{\meanstd{-0.01}{0.02}} & \textcolor{blue}{\textbf{\meanstd{0.51}{0.31}}} \\
& Cost $\downarrow$ & \textcolor{gray}{\meanstd{1.61}{1.53}} & \textbf{\meanstd{0.90}{1.07}} & \textbf{\meanstd{0.00}{0.00}} & \textcolor{blue}{\textbf{\meanstd{0.20}{0.35}}} \\
\midrule
\multirow{2}{*}{SwimmerVelocity} 
& Reward $\uparrow$ & \textcolor{gray}{\meanstd{0.38}{0.10}} & \textbf{\meanstd{-0.02}{0.05}} & \textcolor{blue}{\textbf{\meanstd{0.06}{0.25}}} & {\textbf{\meanstd{0.05}{0.02}}} \\
& Cost $\downarrow$ & \textcolor{gray}{\meanstd{2.11}{3.01}} & \textbf{\meanstd{0.23}{0.20}} & \textcolor{blue}{\textbf{\meanstd{0.91}{1.57}}} & {\textbf{\meanstd{0.00}{0.01}}} \\
\midrule
\multirow{2}{*}{Walker2dVelocity} 
& Reward $\uparrow$ & \textcolor{blue}{\textbf{\meanstd{0.78}{0.01}}} & \textcolor{gray}{\meanstd{0.51}{0.03}} & \textcolor{gray}{\meanstd{0.40}{0.17}} & {\textbf{\meanstd{0.77}{0.02}}} \\
& Cost $\downarrow$ & \textcolor{blue}{\textbf{\meanstd{0.08}{0.07}}} & \textcolor{gray}{\meanstd{1.92}{0.57}} & \textcolor{gray}{\meanstd{5.20}{1.16}} & {\textbf{\meanstd{0.02}{0.02}}} \\
\bottomrule
\end{tabular}

\label{tab:extended_evals}
\end{table}

\ourmethod satisfies the cost constraint in 6 out of the 7  evaluated tasks, demonstrating strong generalization across environments with varying dynamics and cost structures. The next best-performing method, CAPS, remains safe on 5 out of 7 tasks. In terms of reward, \ourmethod achieves the highest normalized return on three environments and remains competitive on the remaining tasks. For example, O3SRL attains performance close to the best methods on \textit{SwimmerVelocity} (0.05 vs.\ 0.06) and \textit{Walker2dVelocity} (0.77 vs.\ 0.78) while maintaining safety.

Collectively, these results show that \ourmethod maintains safety under strict cost constraints while achieving competitive or superior performance across diverse safety-critical environments.

\subsection{Alternative Approach: Offline RL with Fixed $\lambda$ as a Hyperparameter}

{ To assess the effect of treating the Lagrange multiplier $\lambda$ as a fixed hyperparameter, we conducted an additional experiment in which $\lambda$ was held constant throughout training, and the resulting policy was evaluated at the end. This setup departs from the intended offline RL formulation, as choosing an appropriate $\lambda$ would typically require access to online validation data or environment rollouts. Moreover, an exhaustive search over possible $\lambda$ values is computationally impractical and inconsistent with the offline setting.

In our proposed approach (O3SRL), we instead define a discrete set of candidate $\lambda$ values, each corresponding to an ``arm'' in a multi-armed bandit framework. During training, a $\lambda$ value is sampled from a learned distribution over these arms and used to shape the reward. The sampling distribution is updated at every batch following the EXP3 algorithm, enabling adaptive reweighting based on the observed performance of each arm. This mechanism allows the policy to dynamically adjust the effective penalty strength, balancing constraint satisfaction and reward maximization throughout training rather than committing to a single fixed $\lambda$.

Empirically, as shown in Table~\ref{tab:fixed_ablation} fixed-$\lambda$ policies tend to either ignore the cost constraints when $\lambda$ is too small or over-penalize and sacrifice reward when $\lambda$ is too large. By contrast, O3SRL adapts its choice of $\lambda$ during offline training, achieving a more favorable trade-off between reward and cost across different constraint budgets.}

\begin{table}[htbp]
\centering
\caption{DroneCircle results with fixed arms under varying cost limits. The $\uparrow$ symbol denotes that higher rewards are better; $\downarrow$  denotes that lower costs (up to threshold 1) are better. \textbf{Bold}: Safe agents whose normalized cost $\leq$ 1. \textcolor{gray}{Gray}: Unsafe agents. \textbf{\textcolor{blue}{Blue}}: Safe agents with the highest reward.}
% \small
\begin{adjustbox}{max width=0.8\textwidth}
\begin{tabular}{ll|cccccc}
\toprule
\textbf{Cost Limit} &  & Arm 0 & Arm 1 & Arm 2 & Arm 3 & Arm 4 & O3SRL \\
\midrule
\multirow{2}{*}{5} 
& Reward $\uparrow$ & \textcolor{gray}{0.90} & \textcolor{gray}{0.86} & \textcolor{gray}{0.55} & \textbf{0.47} & \textbf{0.39} & \textcolor{blue}{\textbf{0.49}} \\
& Cost $\downarrow$ & \textcolor{gray}{18.18} & \textcolor{gray}{16.52} & \textcolor{gray}{2.47} & \textbf{0.00} & \textbf{0.00} & \textcolor{blue}{\textbf{0.23}} \\
\midrule
\multirow{2}{*}{20} 
& Reward $\uparrow$ & \textcolor{gray}{0.90} & \textcolor{gray}{0.87} & \textcolor{gray}{0.75} & \textbf{0.49} & \textbf{0.40} & \textcolor{blue}{\textbf{0.53}} \\
& Cost $\downarrow$ & \textcolor{gray}{4.54} & \textcolor{gray}{4.33} & \textcolor{gray}{2.80} & \textbf{0.04} & \textbf{0.00} & \textcolor{blue}{\textbf{0.36}} \\
\midrule
\multirow{2}{*}{40} 
& Reward $\uparrow$ & \textcolor{gray}{0.90} & \textcolor{gray}{0.87} & \textcolor{gray}{0.82} & \textbf{0.52} & \textbf{0.42} & \textcolor{blue}{\textbf{0.65}} \\
& Cost $\downarrow$ & \textcolor{gray}{2.27} & \textcolor{gray}{2.21} & \textcolor{gray}{1.88} & \textbf{0.08} & \textbf{0.00} & \textcolor{blue}{\textbf{0.85}} \\
\bottomrule
\end{tabular}
\label{tab:fixed_ablation}
\end{adjustbox}
\end{table}

\subsection{Search space of Lagrange variable $\lambda$}
Recall that our search space for $\lambda$ is $[0, C]$. Table~\ref{tab:C_ablation} presents an ablation study on the impact of varying $C$, the maximum allowable value for the Lagrange variable $\lambda$, which determines how strongly constraint violations are penalized in \ourmethod. Results are shown for $C = 2$, $5$, and $10$. 

The ablation results highlight that small values of $C$ (e.g., $C = 2$) often lead to insufficient constraint enforcement, resulting in catastrophic safety violations in some environments. For example, in \textit{CarCircle} and \textit{DroneCircle}, the average cost exceeds the threshold by a factor of over 8 and 13 respectively, completely defeating the purpose of safe learning. These cases show that when $\lambda$ is capped too low, the algorithm lacks the leverage to penalize unsafe behavior adequately.

In contrast, increasing $C$ to 5 or 10 significantly improves safety across all tasks. For instance, in \textit{DroneCircle}, the cost drops from $13.32$ at $C=2$ to $0.23$ at $C=5$, and further to $0.00$ at $C=10$.  However, in tasks with already low costs (e.g., \textit{AntCircle}), large $C$ values can leads to reduced reward, due to over-penalization. 

Overall, $C = 5$ emerges as a robust choice, offering strong safety guarantees while maintaining high reward across environments. This ablation underscores the necessity of a sufficiently expressive $\lambda$ range to enable \ourmethod to satisfy constraints in practice.

\begin{table}[htbp]
\centering
\caption{\ourmethod results for different values of $C$: $\lambda$ maximum value. Each value shows the average reward ($\uparrow$) and cost ($\downarrow$). Higher reward is better, and lower cost (up to threshold 1) is better. \textbf{Bold}: Safe agents whose normalized cost $\leq$ 1. \textcolor{gray}{Gray}: Unsafe agents.}
\begin{adjustbox}{width=0.7\textwidth}
\begin{tabular}{ll|ccc}
\toprule
\textbf{Tasks} &  & $C=2$ & $C=5$ & $C=10$ \\
\midrule
\multirow{2}{*}{BallRun} & Reward $\uparrow$ & \textbf{\meanstd{0.28}{0.01}} & \textbf{\meanstd{0.25}{0.03}} & \textbf{\meanstd{0.25}{0.01}} \\
& Cost $\downarrow$ & \textbf{\meanstd{0.00}{0.00}} & \textbf{\meanstd{0.00}{0.00}} & \textbf{\meanstd{0.00}{0.00}} \\
\midrule

\multirow{2}{*}{CarRun} & Reward $\uparrow$ & \textbf{\meanstd{0.97}{0.00}} & \textbf{\meanstd{0.96}{0.01}} & \textbf{\meanstd{0.95}{0.01}} \\
& Cost $\downarrow$ & \textbf{\meanstd{0.06}{0.10}} & \textbf{\meanstd{0.02}{0.03}} & \textbf{\meanstd{0.00}{0.00}} \\
\midrule

\multirow{2}{*}{DroneRun} & Reward $\uparrow$ & \textcolor{gray}{\meanstd{0.43}{0.10}} & \textbf{\meanstd{0.32}{0.05}} & \textbf{\meanstd{0.33}{0.05}} \\
& Cost $\downarrow$ & \textcolor{gray}{\meanstd{1.93}{1.90}} & \textbf{\meanstd{0.68}{1.18}} & \textbf{\meanstd{0.38}{0.66}} \\
\midrule

\multirow{2}{*}{AntRun} & Reward $\uparrow$ & \textcolor{gray}{\meanstd{0.34}{0.09}} & \textbf{\meanstd{0.33}{0.13}} & \textbf{\meanstd{0.20}{0.06}} \\
& Cost $\downarrow$ & \textcolor{gray}{\meanstd{1.16}{0.95}} & \textbf{\meanstd{0.14}{0.10}} & \textbf{\meanstd{0.27}{0.47}} \\
\midrule

\multirow{2}{*}{BallCircle} & Reward $\uparrow$ & \textcolor{gray}{\meanstd{0.82}{0.06}} & \textbf{\meanstd{0.62}{0.01}} & \textbf{\meanstd{0.58}{0.03}} \\
& Cost $\downarrow$ & \textcolor{gray}{\meanstd{7.20}{2.99}} & \textbf{\meanstd{0.06}{0.07}} & \textbf{\meanstd{0.01}{0.01}} \\
\midrule

\multirow{2}{*}{CarCircle} & Reward $\uparrow$ & \textcolor{gray}{\meanstd{0.75}{0.11}} & \textbf{\meanstd{0.66}{0.03}} & \textbf{\meanstd{0.57}{0.08}} \\
& Cost $\downarrow$ & \textcolor{gray}{\meanstd{8.07}{7.53}} & \textbf{\meanstd{0.11}{0.16}} & \textbf{\meanstd{0.00}{0.00}} \\
\midrule

\multirow{2}{*}{DroneCircle} & Reward $\uparrow$ & \textcolor{gray}{\meanstd{0.79}{0.06}} & \textbf{\meanstd{0.49}{0.07}} & \textbf{\meanstd{0.37}{0.03}} \\
& Cost $\downarrow$ & \textcolor{gray}{\meanstd{13.32}{2.62}} & \textbf{\meanstd{0.23}{0.34}} & \textbf{\meanstd{0.00}{0.00}} \\
\midrule

\multirow{2}{*}{AntCircle} & Reward $\uparrow$ & \textbf{\meanstd{0.58}{0.02}} & \textbf{\meanstd{0.48}{0.06}} & \textbf{\meanstd{0.41}{0.02}} \\
& Cost $\downarrow$ & \textbf{\meanstd{0.25}{0.26}} & \textbf{\meanstd{0.00}{0.00}} & \textbf{\meanstd{0.04}{0.07}} \\
\bottomrule
\end{tabular}
\end{adjustbox}
\label{tab:C_ablation}
\end{table}

\subsection{Discretization of $\lambda$}
Table~\ref{tab:grid_ablation} presents an ablation comparing two $\lambda$ set choices used in O3SRL: a uniformly spaced set (e.g., $[0, 1.25, 2.5, 3.75, 5]$) and an adaptive set that retains the extreme values 0 and 5 but distributes intermediate values more densely near lower penalty regions (e.g., $[0, 0.5, 1.12, 2.5, 5]$). The purpose of the adaptive grid is to modulate constraint penalization based on the cost threshold $\kappa$, allowing the algorithm to apply smaller penalties in more permissive cost budget scenarios. The results show that while both grids perform similarly under strict cost threshold constraints ($\kappa = 5$), the adaptive grid provides clear advantages as the constraint becomes more relaxed. For instance, under $\kappa = 40$, the adaptive grid allows significantly higher reward in tasks such as \textit{DroneCircle} (reward improves from $0.45$ to $0.65$), by leveraging a smoother trade-off between reward maximization and cost constraint satisfaction. These findings suggest that finer resolution at low penalty levels enables better policy selection for higher cost constraint thresholds, making the adaptive grid more effective for balancing performance and safety.

\begin{table}[htbp]
\centering
\caption{\ourmethod results for different values of the set of $\lambda$s and cost limits $\kappa$. Each value shows the average reward ($\uparrow$) and cost ($\downarrow$). Higher reward is better, and lower cost (up to threshold 1) is better. \textbf{Bold}: Safe agents whose normalized cost $\leq$ 1. \textcolor{gray}{Gray}: Unsafe agents.}
\begin{adjustbox}{width=\textwidth}
\begin{tabular}{ll|cc|cc|cc}
\toprule
\textbf{Tasks} & & \multicolumn{2}{c|}{$\kappa = 5$} & \multicolumn{2}{c|}{$\kappa = 20$} & \multicolumn{2}{c}{$\kappa = 40$} \\
\cmidrule(lr){3-4} \cmidrule(lr){5-6} \cmidrule(lr){7-8}
& & uniform & adaptive & uniform & adaptive & uniform & adaptive \\
\midrule

\multirow{2}{*}{BallRun} & Reward $\uparrow$ & \textbf{\meanstd{0.25}{0.01}} & \textbf{\meanstd{0.25}{0.03}} & \textbf{\meanstd{0.25}{0.01}} & \textbf{\meanstd{0.26}{0.01}} & \textbf{\meanstd{0.25}{0.01}} & \textbf{\meanstd{0.53}{0.48}} \\
& Cost $\downarrow$ & \textbf{\meanstd{0.00}{0.00}} & \textbf{\meanstd{0.00}{0.00}} & \textbf{\meanstd{0.00}{0.00}} & \textbf{\meanstd{0.06}{0.11}} & \textbf{\meanstd{0.00}{0.00}} & \textbf{\meanstd{0.71}{1.24}} \\
\midrule

\multirow{2}{*}{DroneRun} & Reward $\uparrow$ & \textcolor{gray}{\meanstd{0.39}{0.04}} & \textbf{\meanstd{0.32}{0.05}} & \textbf{\meanstd{0.31}{0.11}} & \textbf{\meanstd{0.38}{0.05}} & \textbf{\meanstd{0.30}{0.05}} & \textbf{\meanstd{0.38}{0.10}} \\
& Cost $\downarrow$ & \textcolor{gray}{\meanstd{1.95}{1.97}} & \textbf{\meanstd{0.68}{1.18}} & \textbf{\meanstd{0.22}{0.33}} & \textbf{\meanstd{0.41}{0.41}} & \textbf{\meanstd{0.60}{0.72}} & \textbf{\meanstd{0.44}{0.44}} \\
\midrule

\multirow{2}{*}{BallCircle} & Reward $\uparrow$ & \textbf{\meanstd{0.63}{0.04}} & \textbf{\meanstd{0.62}{0.01}} & \textbf{\meanstd{0.63}{0.08}} & \textbf{\meanstd{0.69}{0.09}} & \textbf{\meanstd{0.62}{0.06}} & \textbf{\meanstd{0.76}{0.04}} \\
& Cost $\downarrow$ & \textbf{\meanstd{0.07}{0.12}} & \textbf{\meanstd{0.06}{0.07}} & \textbf{\meanstd{0.20}{0.30}} & \textbf{\meanstd{0.58}{0.66}} & \textbf{\meanstd{0.10}{0.10}} & \textbf{\meanstd{0.64}{0.18}} \\
\midrule

\multirow{2}{*}{DroneCircle} & Reward $\uparrow$ & \textbf{\meanstd{0.46}{0.02}} & \textbf{\meanstd{0.49}{0.07}} & \textbf{\meanstd{0.46}{0.01}} & \textbf{\meanstd{0.53}{0.01}} & \textbf{\meanstd{0.45}{0.03}} & \textbf{\meanstd{0.65}{0.11}} \\
& Cost $\downarrow$ & \textbf{\meanstd{0.00}{0.00}} & \textbf{\meanstd{0.23}{0.34}} & \textbf{\meanstd{0.00}{0.00}} & \textbf{\meanstd{0.36}{0.27}} & \textbf{\meanstd{0.00}{0.00}} & \textbf{\meanstd{0.85}{0.55}} \\

\bottomrule
\end{tabular}
\end{adjustbox}

\label{tab:grid_ablation}
\end{table}

\subsection{Number of update steps $M$}

Recall that our approximate algorithm for empirical evaluation performs $M$ stochastic gradient updates of a given offline RL algorithm in each iteration. 
Table \ref{tab:M_steps_ablation} presents an ablation on the number of offline RL update steps $M$. This parameter controls how long the policy is trained before updating the $\lambda$ value to balance reward and constraint satisfaction. The results demonstrate that $M = 10$ provides a stable trade-off between learning progress and responsiveness to constraint violations. With $M = 1$, the frequent updates to $\lambda$ overall increases costs adjusting the policy aggressively. Conversely, setting $M = 100$ delays $\lambda$ updates, which can result in policies drifting away from feasible regions, as seen in the sharp increase in costs for DroneRun and DroneCircle. Overall, the ablation results validate that safe and effective offline learning in \ourmethod requires coordinated progress on both the policy and the constraint via well-timed $\lambda$ updates. The choice of $M = 10$ strikes this balance, supporting its use for all the results presented in the main paper.

\begin{table}[htbp]
\centering
\caption{\ourmethod results for different values of $M$: Offline RL update steps. Each value shows the average reward ($\uparrow$) and cost ($\downarrow$). Higher reward is better, and lower cost (up to threshold 1) is better. \textbf{Bold}: Safe agents whose normalized cost $\leq$ 1. \textcolor{gray}{Gray}: Unsafe agents.}
\small
\begin{adjustbox}{width=0.7\textwidth}
\begin{tabular}{ll|ccc}
\toprule
\textbf{Tasks} &  & $M=1$ & $M=10$ & $M=100$ \\
\midrule
\multirow{2}{*}{BallRun} & Reward $\uparrow$ & \textbf{\meanstd{0.25}{0.03}} & \textbf{\meanstd{0.25}{0.03}} & \textbf{\meanstd{0.26}{0.03}} \\
& Cost $\downarrow$ & \textbf{\meanstd{0.37}{0.64}} & \textbf{\meanstd{0.00}{0.00}} & \textbf{\meanstd{0.00}{0.00}} \\
\midrule

\multirow{2}{*}{CarRun} & Reward $\uparrow$ & \textbf{\meanstd{0.96}{0.00}} & \textbf{\meanstd{0.96}{0.01}} & \textbf{\meanstd{0.96}{0.00}} \\
& Cost $\downarrow$ & \textbf{\meanstd{0.00}{0.01}} & \textbf{\meanstd{0.02}{0.03}} & \textbf{\meanstd{0.01}{0.01}} \\
\midrule

\multirow{2}{*}{DroneRun} & Reward $\uparrow$ & \textcolor{gray}{\meanstd{0.34}{0.06}} & \textbf{\meanstd{0.32}{0.05}} & \textcolor{gray}{\meanstd{0.27}{0.08}} \\
& Cost $\downarrow$ & \textcolor{gray}{\meanstd{1.10}{0.90}} & \textbf{\meanstd{0.68}{1.18}} & \textcolor{gray}{\meanstd{3.72}{4.44}} \\
\midrule

\multirow{2}{*}{AntRun} & Reward $\uparrow$ & \textbf{\meanstd{0.34}{0.05}} & \textbf{\meanstd{0.33}{0.13}} & \textbf{\meanstd{0.39}{0.16}} \\
& Cost $\downarrow$ & \textbf{\meanstd{0.20}{0.21}} & \textbf{\meanstd{0.14}{0.10}} & \textbf{\meanstd{0.97}{1.36}} \\
\midrule

\multirow{2}{*}{BallCircle} & Reward $\uparrow$ & \textbf{\meanstd{0.67}{0.00}} & \textbf{\meanstd{0.62}{0.01}} & \textbf{\meanstd{0.67}{0.05}} \\
& Cost $\downarrow$ & \textbf{\meanstd{0.25}{0.19}} & \textbf{\meanstd{0.06}{0.07}} & \textbf{\meanstd{0.61}{0.88}} \\
\midrule

\multirow{2}{*}{CarCircle} & Reward $\uparrow$ & \textbf{\meanstd{0.66}{0.01}} & \textbf{\meanstd{0.66}{0.03}} & \textbf{\meanstd{0.68}{0.01}} \\
& Cost $\downarrow$ & \textbf{\meanstd{0.10}{0.17}} & \textbf{\meanstd{0.11}{0.16}} & \textbf{\meanstd{0.17}{0.29}} \\
\midrule

\multirow{2}{*}{DroneCircle} & Reward $\uparrow$ & \textbf{\meanstd{0.44}{0.06}} & \textbf{\meanstd{0.49}{0.07}} & \textcolor{gray}{\meanstd{0.58}{0.20}} \\
& Cost $\downarrow$ & \textbf{\meanstd{0.11}{0.10}} & \textbf{\meanstd{0.23}{0.34}} & \textcolor{gray}{\meanstd{4.87}{7.77}} \\
\midrule

\multirow{2}{*}{AntCircle} & Reward $\uparrow$ & \textbf{\meanstd{0.41}{0.06}} & \textbf{\meanstd{0.48}{0.06}} & \textbf{\meanstd{0.49}{0.01}} \\
& Cost $\downarrow$ & \textbf{\meanstd{0.02}{0.03}} & \textbf{\meanstd{0.00}{0.00}} & \textbf{\meanstd{0.00}{0.00}} \\
\bottomrule
\end{tabular}
\end{adjustbox}

\label{tab:M_steps_ablation}
\end{table}

\subsection{Evaluation Across Multiple Random Seeds}

{To assess the stability of \ourmethod with respect to random initialization, we extend our evaluation to include additional random seeds. Specifically, we compare results averaged over the original three seeds (10, 20, and 30) with those averaged over all six seeds (10, 20, 30, 40, 50, and 60). Table \ref{tab:seed_comparison} reports the mean and standard deviation of rewards and costs under both settings.

The results across confirm the stability of \ourmethod, showing consistent rewards and safe cost levels across all evaluated environments. The performance trends observed with three seeds largely persist when extended to six, indicating robust and reliable behavior under different random seeds.}

\begin{table}[htbp]
\centering
\caption{Comparison between 3-seed and 6-seed experiments. Reward ($\uparrow$) indicates higher is better, and Cost ($\downarrow$) indicates lower is better. \textbf{Bold}: Safe agents whose normalized cost $\leq$ 1.}
\small
\begin{tabular}{ll|cc}
\toprule
\textbf{Environment} &  & 3 seeds & 6 seeds  \\
\midrule
\multirow{2}{*}{BallRun} 
& Reward $\uparrow$ & \textbf{\meanstd{0.25}{0.03}} & {\textbf{\meanstd{0.26}{0.02}}}   \\
& Cost $\downarrow$ &  \textbf{\meanstd{0.00}{0.00}} &   {\textbf{\meanstd{0.00}{0.00}}} \\
\midrule
\multirow{2}{*}{CarRun} 
& Reward $\uparrow$ & \textbf{\meanstd{0.96}{0.01}} & {\textbf{\meanstd{0.96}{0.01}}}   \\
& Cost $\downarrow$ &  \textbf{\meanstd{0.02}{0.03}} & {\textbf{\meanstd{0.00}{0.00}}} \\
\midrule
\multirow{2}{*}{DroneRun} 
& Reward $\uparrow$ & \textbf{\meanstd{0.32}{0.05}} & {\textbf{\meanstd{0.37}{0.03}}}   \\
& Cost $\downarrow$ &  \textbf{\meanstd{0.68}{1.18}} & {\textbf{\meanstd{0.74}{0.67}}} \\
\midrule
\multirow{2}{*}{AntRun} 
& Reward $\uparrow$ & \textbf{\meanstd{0.33}{0.13}} & {\textbf{\meanstd{0.22}{0.10}}}  \\
& Cost $\downarrow$ & \textbf{\meanstd{0.14}{0.10}} & {\textbf{\meanstd{0.17}{0.19}}} \\
\midrule
\multirow{2}{*}{BallCircle} 
& Reward $\uparrow$ & \textbf{\meanstd{0.62}{0.01}} & {\textbf{\meanstd{0.63}{0.04}}}   \\
& Cost $\downarrow$ & \textbf{\meanstd{0.06}{0.07}} & {\textbf{\meanstd{0.26}{0.25}}} \\
\midrule
\multirow{2}{*}{CarCircle} 
& Reward $\uparrow$ & \textbf{\meanstd{0.66}{0.03}} & {\textbf{\meanstd{0.66}{0.03}}}   \\
& Cost $\downarrow$ & \textbf{\meanstd{0.11}{0.16}} & {\textbf{\meanstd{0.04}{0.10}}} \\
\midrule
\multirow{2}{*}{DroneCircle} 
& Reward $\uparrow$ & \textbf{\meanstd{0.49}{0.07}} & {\textbf{\meanstd{0.47}{0.03}}}  \\
& Cost $\downarrow$ & \textbf{\meanstd{0.23}{0.34}} & {\textbf{\meanstd{0.02}{0.04}}} \\
\midrule
\multirow{2}{*}{AntCircle} 
& Reward $\uparrow$ & \textbf{\meanstd{0.48}{0.06}} & {\textbf{\meanstd{0.44}{0.05}}} \\
& Cost $\downarrow$ & \textbf{\meanstd{0.00}{0.00}} & {\textbf{\meanstd{0.02}{0.03}}} \\
\bottomrule
\end{tabular}
\label{tab:seed_comparison}
\end{table}

\section{Experimental Details}

This section provides additional details about the experimental setup and hyper-parameters.

\subsection{Description of environments}
\begin{figure}[htbp]
    \centering
    \includegraphics[width=0.9\textwidth]{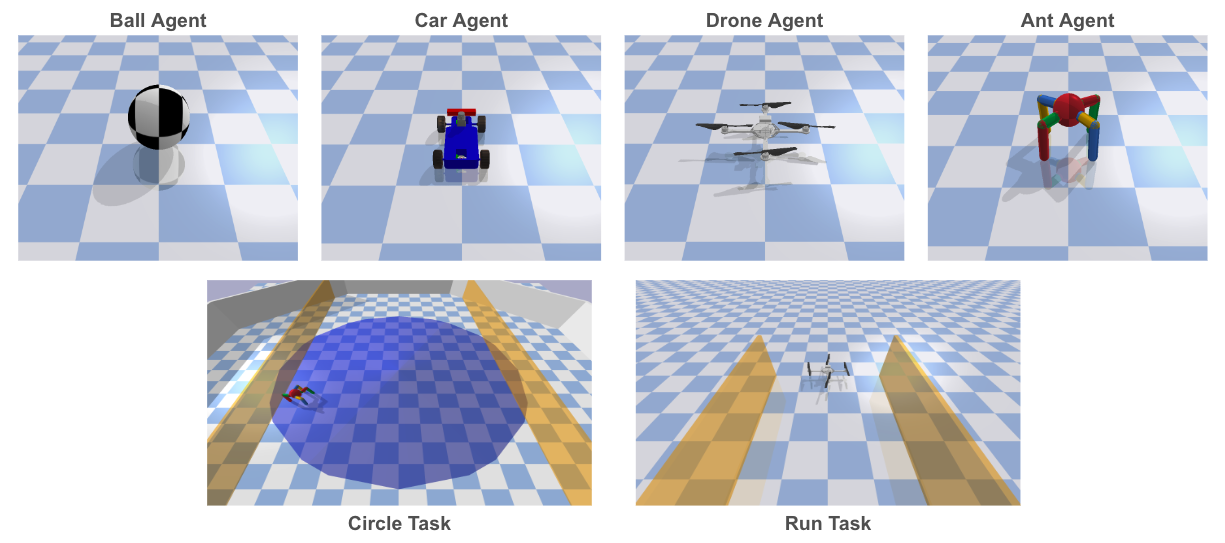} % Make sure the file is named pybullet.[extension]
    \caption{Visualization of the Bullet-Safety-Gym environments featuring different agents and tasks.}
    \label{fig:pybullet}
\end{figure}

We evaluate \ourmethod performance in a set of environments from the \textbf{Bullet-Safety-Gym} suite. Each environment is defined by a combination of \textit{robot type} and \textit{task}, resulting in scenarios such as \textit{BallRun}, \textit{CarCircle}, \textit{DroneRun}, and so on. The agent types include \textbf{Ball}, \textbf{Car}, \textbf{Drone}, and \textbf{Ant}, and each is evaluated under either the \textbf{Run} or \textbf{Circle} task. Figure \ref{fig:pybullet} illustrates the Bullet-Safety-Gym environments, demonstrating various agent types (Ball, Car, Drone, Ant) performing the Run and Circle tasks.

\paragraph{Run Environments:} In the Run task, agents are required to move quickly along a straight corridor. They receive rewards based on forward progress, encouraging high-speed locomotion. However, they are penalized for crossing lateral safety boundaries or exceeding a predefined velocity threshold. These safety constraints are not enforced through physical barriers but through a cost signal that tracks violations.

\paragraph{Circle Environments:} In the Circle task, agents are trained to move in a clockwise direction around a circular track. Rewards are higher when agents maintain fast, smooth motion near the boundary of the circle. Deviating from the intended path or exiting the safety zone results in penalties.

\subsection{Training details}
\paragraph{Adaptive $\lambda$ set:} We construct the $\lambda$ set using a geometric spacing scheme that adapts to the cost limit. The grid always includes $\lambda = 0$ and a maximum value $\lambda_{\text{max}}=C$, and places $k - 2$ intermediate points between them. These intermediate values are spaced geometrically and scaled by a factor $\text{shrink} = (\text{reference\_limit} / \text{cost\_limit})^{\alpha_{\text{shrink}}}$, which compresses the grid as the cost limit increases, where $\text{reference\_limit} = 5$ and $\alpha_{\text{shrink}} = 0.3$. This design ensures finer resolution among low-penalty values when constraints are loose, and broader coverage when strict cost constraints require stronger penalization. 

\begin{figure}[htbp]
    \centering
    \includegraphics[width=0.5\textwidth]{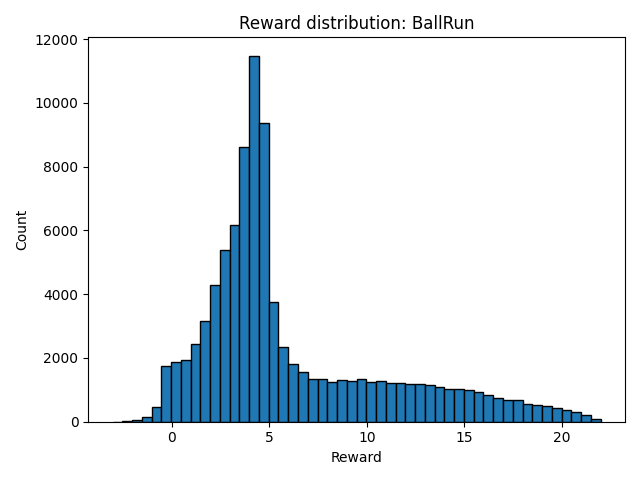} % adjust filename as needed
    \caption{An example of a heavy-tailed reward distribution from the \textit{BallRun} dataset.}
    \label{fig:reward_tail_example}
\end{figure}

\paragraph{Rewards scale:} 
We observe that reward distributions in offline datasets can be heavy-tailed, with occasional extreme values that disproportionately affect learning stability, as illustrated in Figure~\ref{fig:reward_tail_example}.

To mitigate this, we clip rewards symmetrically at the $\tau$-th percentile of their absolute values:
\[
r_{\text{clip}} = \text{percentile}(|r|, \tau), \quad r \leftarrow \text{clip}(r, -r_{\text{clip}}, r_{\text{clip}})
\]
where $\tau =99$. We then scale down the clipped rewards multiplying by $0.9 /r_\text{clip}$.

\paragraph{Hyperparameters:} We employ a batch size of 1024 for the tasks \textit{CarCircle}, \textit{DroneRun}, \textit{AntCircle}, and \textit{BallCircle}, while a batch size of 512 is used for the remaining tasks: \textit{CarRun}, \textit{DroneCircle}, \textit{AntRun}, and \textit{BallRun}. The other hyperparameters used in our experiments are listed in Table \ref{tab:hyperparameters}. For TD3+BC, we adopt the same hyperparameter settings as in the original paper.

\begin{table*}[ht]
\caption{\ourmethod Hyperparameters}
\centering
\begin{tabular}{lc}
\midrule
\textbf{ EXP3 parameters} &  \\ 
\midrule
Number of arms $k$ & 5 \\
$C = \max(\lambda)$ & 5 \\
$\lambda$ update frequency $M$ & 10 \\
$\lambda$ learning rate $\eta$ & 2e-3 \\
\midrule
\textbf{TD3BC parameters} &  \\ 
\midrule
Discount $\gamma$ & 0.99 \\
Policy noise & 0.2\\
Policy noise clip & (0.5, 0.5)\\
Policy update frequency & 2 \\
$\alpha$ & 2.5\\
Optimizer & Adam \\ 
Actor, Critic learning rate & 3e-4 \\
Actor, Critic hidden size & 256  \\ 
Training steps & 100000 \\ 
\midrule
Seed & [10, 20, 30] \\
\midrule
\end{tabular}
\label{tab:hyperparameters}
\end{table*}

\begin{figure}[htbp]
    \centering
    \includegraphics[width=\textwidth]{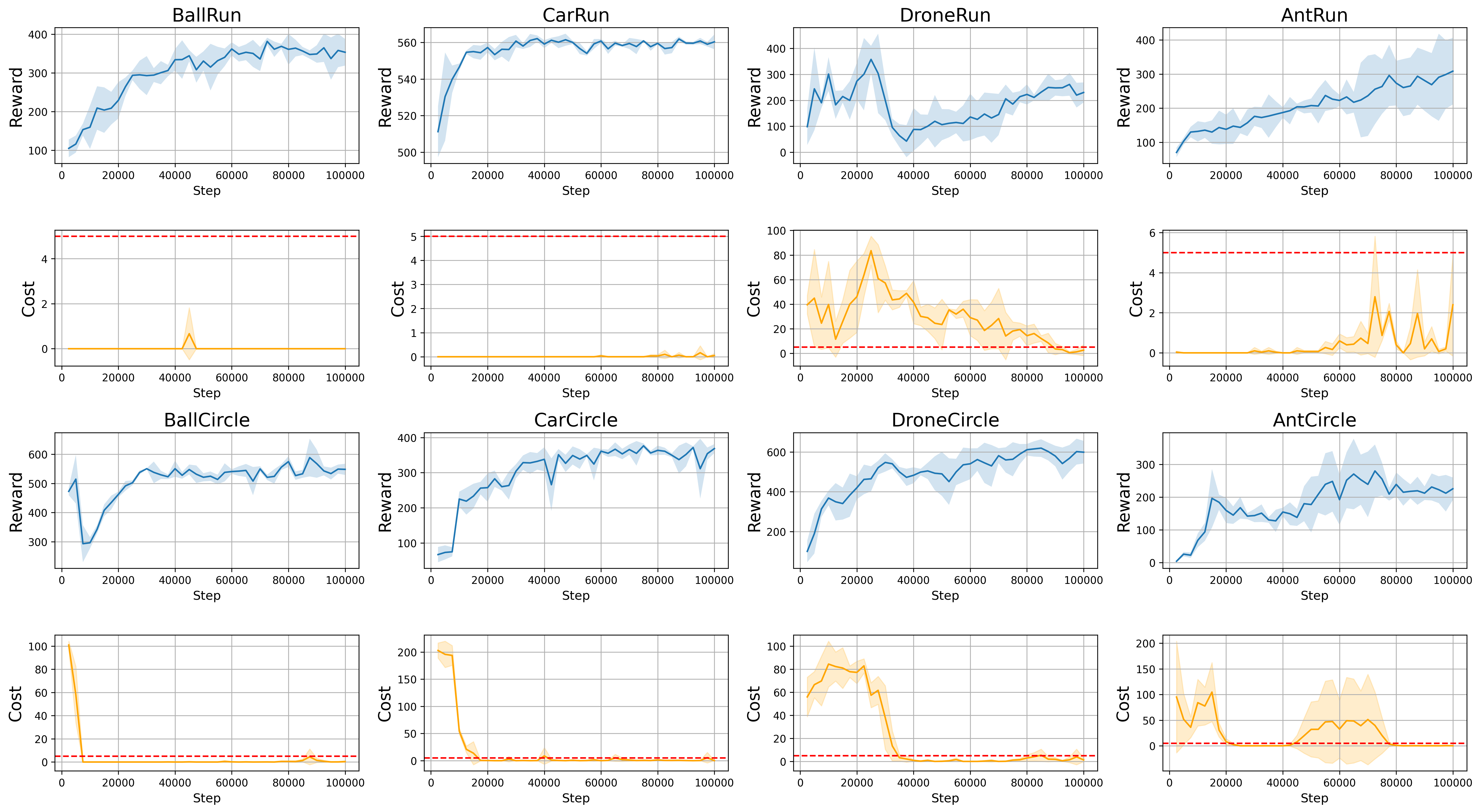} 
    \caption{Training curves of \ourmethod across different tasks, showing average reward (top) and cost (bottom) over training iterations.}
    \label{fig:o3srl_training_curves}
\end{figure}

\paragraph{Training Curves of \ourmethod:} 
Figure \ref{fig:o3srl_training_curves} shows the training curves of \ourmethod across the different tasks, plotting both the average reward and constraint cost over training iterations. The curves demonstrate that \ourmethod is able to quickly stabilize and achieve a favorable trade-off between reward maximization and cost minimization. In most tasks, the cost decreases steadily and remains below the constraint threshold, while the reward improves progressively.

\paragraph{Computational Resources and Training Time:}
All experiments were conducted using an NVIDIA A40 GPU with 48GB of memory. Training a single task for one random seed takes approximately 20 minutes when the server is not under load.